\documentclass[lettersize,journal]{IEEEtran}

\usepackage{microtype}
\usepackage{graphicx}
\usepackage{booktabs} 
\usepackage{hyperref}
\usepackage{multirow}

\usepackage{amsmath}
\usepackage{amssymb}
\usepackage{mathtools}
\usepackage{amsthm}

\usepackage{dsfont}

\usepackage[capitalize,noabbrev]{cleveref}

\usepackage{comment}

\theoremstyle{plain}
\newtheorem{theorem}{Theorem}[section]

\newtheorem{corollary}[theorem]{Corollary}
\newtheorem{innercustomthm}{Theorem}
\newenvironment{appthm}[1]
{\renewcommand\theinnercustomthm{#1}\innercustomthm}
  {\endinnercustomthm}
\theoremstyle{definition}
\newtheorem{definition}[theorem]{Definition}

\theoremstyle{remark}

\usepackage[textsize=tiny]{todonotes}

\DeclareMathOperator{\Prob}{\mathbb{P}}
\DeclareMathOperator{\E}{\mathbb{E}}
\DeclareMathOperator*{\argmax}{arg\,max}

\begin{document}

\title{Robust Classification with Noisy Labels Based on Posterior Maximization}

\author{Nicola Novello and Andrea M. Tonello
\thanks{The authors are with the Institute of Networked and Embedded Systems, University of Klagenfurt, Klagenfurt, Austria (e-mail: \{nicola.novello, andrea.tonello\}@aau.at).}
}


\maketitle

\begin{abstract}
Designing objective functions robust to label noise is crucial for real-world classification algorithms. In this paper, we investigate the robustness to label noise of an $f$-divergence-based class of objective functions recently proposed for supervised classification, herein referred to as $f$-PML. We show that, in the presence of label noise, any of the $f$-PML objective functions can be corrected to obtain a neural network that is equal to the one learned with the clean dataset. Additionally, we propose an alternative and novel correction approach that, during the test phase, refines the posterior estimated by the neural network trained in the presence of label noise. Then, we demonstrate that, even if the considered $f$-PML objective functions are not symmetric, they are robust to symmetric label noise for any choice of $f$-divergence, without the need for any correction approach. This allows us to prove that the cross-entropy, which belongs to the $f$-PML class, is robust to symmetric label noise. Finally, we show that such a class of objective functions can be used together with refined training strategies, achieving competitive performance against state-of-the-art techniques of classification with label noise.
\end{abstract}


\section{Introduction}
\label{sec:introduction}

The success of large deep neural networks is highly dependent on the availability of large labeled datasets. However, the labeling process is often expensive and sometimes imprecise, either if it is done by human operators or by automatic labeling tools. On average, datasets contain from $8\%$ to $38.5\%$ of samples that are corrupted with label noise \cite{song2022learning, xiao2015learning, li2017webvision, lee2018cleannet, song2019selfie}. 

For classification tasks, different lines of research focused on the architecture and training strategy development or on the objective function design.  
For supervised classification tasks, various objective functions have been proposed with the goal of replacing the cross-entropy (CE) \cite{hui2020evaluation, dong2019single, blondel2019learning, pmlr-v235-novello24a}, achieving promising results. 
Meanwhile, in the weakly-supervised scenario of classification with label noise, various evidence showed that the standard CE minimization is not the best option \cite{ghosh2017robust, zhang2018generalized, liu2020early}. 

In this paper, we show that the class of objective functions relying on the maximum a posteriori probability (MAP) approach proposed in \cite{pmlr-v235-novello24a} for supervised classification (referred to as $f$-divergence based Posterior Maximization Learning ($f$-PML) in this paper), is an effective option also in the presence of label noise. 
We propose two correction techniques to make $f$-PML robust to label noise. The first has to be applied during the training phase, similarly to other approaches, to learn a neural network that is equal to the neural network learned with the clean dataset.
For the second, instead, we show that the MAP-based formulation of the classification problem allows to express the posterior in the presence of label noise as a function of the neural network's output and the noise transition probabilities. This allows the design of a novel correction method applied during the test phase to correct the posterior estimate, making it tolerant to label noise.
Moreover, we show that, although $f$-PML objective functions are not symmetric, they are robust to symmetric label noise for \emph{any} $f$-divergence without requiring the estimation of the noise transition probabilities, for mild conditions on the noise rates. As a side (but fundamental) outcome, we demonstrate that the CE is robust to symmetric label noise, correcting many previous papers that affirmed the contrary by relying on the fact that it is not symmetric. 
In addition, we observe that $f$-PML can be seen as a specific case of active passive losses (APLs) \cite{ma2020normalized}, and attains significantly higher accuracy than other APL-like losses. 
Finally, we combine the robust losses with refined training strategies to demonstrate that $f$-PML can also be used with complex training strategies to achieve a competitive performance with state-of-the-art techniques.

The key contributions of this paper are:
\begin{itemize}
    \item We prove the robustness of $f$-PML to symmetric label noise for \emph{any} $f$-divergence, without requiring the class of objective functions to be symmetric. As a key byproduct, we demonstrate the robustness of the CE.
    \item For label noise models where $f$-PML is not robust to label noise, we provide novel approaches to correct either the objective function or the posterior estimator, to achieve robustness.  
    \item Our experimental results show that $f$-PML can be used jointly with refined training strategies to achieve performance competitive with state-of-the-art techniques.
\end{itemize}



\section{Related Work}
\label{sec:related}

In this section, we provide a brief summary of the existing approaches for classification in the presence of label noise. 

\paragraph{Objective function correction} These methods all rely on the idea of modifying the objective function to improve the classifier's label noise robustness. These algorithms require to know the matrix of transition probabilities from true labels to fake labels (i.e., transition matrix). When the transition matrix is not known, it can be estimated, as studied in \cite{patrini2017making, yao2020dual,li2021provably, zhang2021learning, cheng2022class}.
In \cite{natarajan2013learning}, the authors propose a weighted loss function for binary classification in the presence of class-conditional noise.
In \cite{liu2015classification}, the authors utilize the transition matrix to employ reweighting, which utilizes importance sampling to ensure robustness.
Forward and Backward \cite{patrini2017making} are two algorithms for loss correction given the transition matrix, which is estimated finding the dataset anchor points. 
In \cite{ma2018dimensionality}, the authors propose a loss correction approach to avoid the overfitting of noisy labels during the dimensionality expansion phase of the training process.
In \cite{bae2024dirichlet}, the authors propose a resampling technique that works better than reweighting in the label noise scenario. Shifted Gaussian Noise (SGN) \cite{englesson2024robust} provides a method combining loss reweighting and label correction.

\paragraph{Robust objective functions} These algorithms utilize objective functions that are theoretically robust to label noise without the need of estimating the transition probabilities. 
In \cite{ghosh2017robust}, the authors prove the robustness of symmetric objective functions. In particular, they show that the CE is not symmetric, while proving that the mean absolute error (MAE) is a robust loss. 
In \cite{zhang2018generalized}, the authors show that MAE performs poorly for challenging datasets and propose the generalized cross entropy (GCE), which is a trade-off between MAE and categorical CE, leveraging the negative Box-Cox transformation. 
Symmetric Cross Entropy (SCE) \cite{wang2019symmetric} combines the CE loss with a Reverse Cross Entropy (RCE) loss robust to label noise, to avoid overfitting to noisy labels.
In \cite{xu2019l_dmi}, the authors propose a robust loss function based on the determinant based mutual information.
In \cite{ma2020normalized}, the authors prove that all the objective functions can be made robust to label noise with a normalization. However, they show that robust losses can tend to underfit. Therefore, they propose a class of objective functions, referred to as active passive losses (APLs), that mitigate the underfitting problem. 
Peer Loss functions \cite{liu2020peer} are a class of robust loss functions inspired by correlated agreement.
In \cite{wei2020optimizing}, the authors propose a class of objective functions based on the maximization of the $f$-divergence-based generalization of mutual information. 
In \cite{ye2023active}, the authors propose a specific class of APLs, referred to as active negative loss functions (ANLs), that, instead of obtaining the passive losses based on MAE as in \cite{ma2020normalized}, use negative loss functions based on complementary label learning \cite{ishida2017learning}.
In \cite{zhou2023asymmetric}, the authors propose a class of loss functions robust to label noise that extend symmetric losses. Furthermore, they highlight the importance of designing objective functions that are not symmetric and robust to label noise.

\paragraph{Refined training strategies} These algorithms rely on elaborated training strategies that improve the robustness to label noise. Many techniques use ensemble models.
MentorNet \cite{jiang2018mentornet} supervises a student network by providing it a data-driven curriculum.
Co-teaching \cite{han2018co} trains two networks simultaneously using the most confident predictions of one network to train the other one. 
For Co-teaching+ \cite{yu2019does}, the authors propose to bridge the Co-teaching and update with disagreement frameworks. \\
Some techniques rely on semi-supervised learning and sample selection techniques. In \cite{berthelot2019mixmatch}, the authors unify many semi-supervised learning approaches in one algorithm. Divide-Mix \cite{li2020dividemix} uses label co-refinement and label co-guessing during the semi-supervised learning phase. In \cite{wang2022promix}, the authors propose an algorithm that uses a new progressive selection technique to select clean samples. 
Contrastive frameworks have also been used in popular approaches. For instance, Joint training with Co-Regularization (JoCoR) \cite{wei2020combating} aims to reduce the diversity of two networks during training, minimizing a contrastive loss. Other contrastive learning-based algorithms are proposed in \cite{ghosh2021contrastive, yi2022learning}. \\
Other techniques rely on gradient clipping \cite{menon2020can}, logit clipping \cite{wei2023mitigating}, label smoothing \cite{wei2022smooth}, regularization \cite{cheng2021learning, liu2020early, liu2022robust, cheng2021mitigating}, meta-learning \cite{li2019learning}, area under the margin statistic \cite{pleiss2020identifying}, data ambiguation \cite{lienen2024mitigating}, early stopping \cite{huang2023paddles, yuan2024early}, and joint optimization of network parameters and data labels \cite{tanaka2018joint}.

\section{Robust $f$-Divergence MAP Classification with Label Noise}
\label{sec:robust_classification}

\begin{figure*}[ht]
	\centering
	\includegraphics[width=\textwidth]{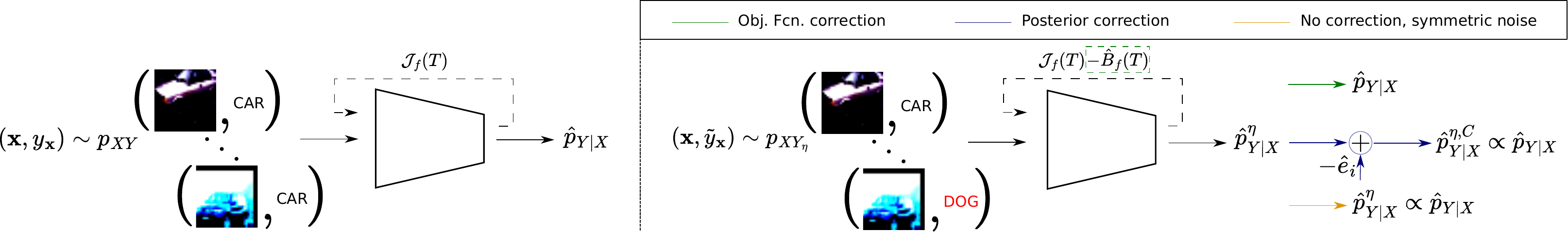}
	\caption{Proposed framework in the absence (left) and presence (right) of label noise. To achieve robustness with label noise, the objective function correction (green) is performed during training to obtain the clean estimate of the posterior as the output of the neural network. Alternatively, the posterior correction (blue) is implemented during test by correcting the posterior estimate. In the case of symmetric noise, the proposed framework does not require any correction technique. The dashed arrows indicate the model update through backpropagation.} \label{fig:framework}
\end{figure*}

In Sec. \ref{subsec:f_div} and \ref{subsec:PMI_classification}, we provide the necessary preliminaries related to the $f$-divergence and the MAP-based supervised classification approach. In Sec. \ref{subsec:obj_fcn_correction} and \ref{subsec:estimator_correction}, we present the novel objective function and posterior correction approaches. In Sec. \ref{subsec:robustness_analysis}, we demonstrate $f$-PML's robustness to symmetric label noise without requiring the knowledge of noise rates. In Sec. \ref{subsec:convergence_analysis}, we study the convergence of $f$-PML in the presence of label noise.  Finally, in Sec. \ref{subsec:comparison_related}, we show intriguing relationships between $f$-PML and part of the related work. The block diagram representing the whole framework is reported in Fig. \ref{fig:framework}.

\subsection{$f$-Divergence}
\label{subsec:f_div}
Given a domain $\mathcal{X}$ and two probability density functions  $p(\mathbf{x})$, $q(\mathbf{x})$ on this domain, the $f$-divergence is defined as \cite{ali1966general, csiszar1967information} 
\begin{equation}
    D_f(p||q) = \int_{\mathcal{X}} q(\mathbf{x})  f\left( \frac{p(\mathbf{x})}{q(\mathbf{x})} \right)   d\mathbf{x}
\end{equation}
where $p \ll q$ (i.e., $p$ is absolutely continuous with respect to $q$) and  where the \textit{generator function} $f: \mathbb{R}_+ \longrightarrow \mathbb{R}$ is a convex, lower-semicontinuous function such that $f(1)=0$.
The variational representation of the $f$-divergence \cite{Nguyen2010} reads as
\begin{equation}
\label{eq:variational_representation}
    D_f(p||q) = \sup_{T: \mathcal{X} \to  \mathbb{R}} \left\{ \E_{p} \left[ T(\mathbf{x}) \right] - \E_{q} \left[ f^*(T(\mathbf{x})) \right] \right\}.
\end{equation}
where $T$ is a parametric function (e.g., a neural network) and $f^*$ denotes the \textit{Fenchel conjugate} of $f$ and is defined as 
\begin{equation}
\label{eq:fenchel_conjugate}
    f^*(t) = \sup_{u \in dom_f} \left\{ ut - f(u) \right\},
\end{equation}
with $dom_f$ being the domain of the function $f$.   
The supremum over all functions in  \eqref{eq:variational_representation} is attained for 
\begin{equation}
\label{eq:T_hat}
    T^\diamond(\mathbf{x}) = f^{\prime} \left( \frac{p(\mathbf{x})}{q(\mathbf{x})} \right) ,
\end{equation}
where $f^\prime$ is the first derivative of $f$. 

\subsection{MAP-Based Classification}
\label{subsec:PMI_classification}

In this section, we highlight an information-theoretic perspective of the MAP approach and introduce the related discriminative MAP-based classification algorithm \cite{pmlr-v235-novello24a}. Then, we provide the necessary preliminaries on classification in the presence of label noise. 

Mutual information (MI) is a statistical quantity that measures the dependency between random vectors. Let $X \in \mathcal{X}$ and $Y \in \mathcal{Y}$ be two random vectors having probability density functions $p_X(\mathbf{x})$ and $p_Y(\mathbf{y})$, respectively. Let $y_\mathbf{x}$ be the label corresponding to an object $\mathbf{x}$ (e.g., an image), the MI between $X$ and $Y$ is defined as 
\begin{align}
    I(X;Y) = \E_{{XY}}\Biggl[ \underbrace{\log\left( \frac{p_{XY}(\mathbf{x}, y_{\mathbf{x}})}{p_X(\mathbf{x})p_Y(y_{\mathbf{x}})} \right)}_{\triangleq \iota(\mathbf{x};y_\mathbf{x})} \Biggr], 
\end{align}
where $\iota(\mathbf{x};y_\mathbf{x})$ is the pointwise mutual information (PMI). Let $\mathcal{A}_y$ be a set of $K$ classes, maximizing the PMI corresponds to solving the MAP classification criterion, i.e., 
\begin{align}
    \hat{y}_\mathbf{x} = \argmax_{y_\mathbf{x} \in \mathcal{A}_y} \iota(\mathbf{x};y_\mathbf{x}) = \argmax_{y_\mathbf{x} \in \mathcal{A}_y} p_{Y|X}(y_\mathbf{x}|\mathbf{x}) ,
\end{align}
because $p_Y$ is fixed given a dataset. 

From \cite{pmlr-v235-novello24a}, leveraging a discriminative formulation that allows to estimate the posterior as a density ratio, the supervised classification problem can be solved by maximizing over the neural network $T(\cdot)$ the objective function
\begin{align}
    \label{eq:supervised_general_value_function}
        \mathcal{J}_f(T) &= \E_{XY}\Biggl[ T(\mathbf{x}) \textbf{1}_K(y_\mathbf{x})\Biggr] - \E_{X}\Biggl[ \sum_{i=1}^K f^{*} \left(T(\mathbf{x},i) \right) \Biggr] ,
    \end{align}
where ${T}(\mathbf{x}) = [T(\mathbf{x},1), \dots, T(\mathbf{x},K)]$, with $T(\mathbf{x},i)$ $i$-th component of the neural network's output $T(\mathbf{x})$, and $\textbf{1}_K(y_\mathbf{x})$ is the one-hot encoded label $y_\mathbf{x}$.
The class to which $\mathbf{x}$ belongs is estimated as
\begin{align}
\label{eq:MAP}
    \hat{y}_\mathbf{x} &= \argmax_{y_\mathbf{x} \in \mathcal{A}_y} p_{Y|X}(y_\mathbf{x}|\mathbf{x}) = \argmax_{y_\mathbf{x} \in \mathcal{A}_y} \> (f^*)^\prime({T}^\diamond(\mathbf{x})) ,
\end{align}
where ${T}^\diamond(\cdot)$ is the optimal neural network trained by maximizing \eqref{eq:supervised_general_value_function}. It should be noted that \eqref{eq:supervised_general_value_function} is actually a class of objective functions each defined for a given choice of $f$.

In a supervised classification problem, the neural network $T$ is learned using a clean dataset $\{(\mathbf{x}_1,y_{\mathbf{x}_1}), \dots, (\mathbf{x}_N, y_{\mathbf{x}_N})\} \equiv \mathcal{D}$ drawn i.i.d. from $X \times Y$.
Differently, in the weakly-supervised scenario of classification in presence of label noise, we can only access a noisy dataset $\{(\mathbf{x}_1,\tilde{y}_{\mathbf{x}_1}), \dots, (\mathbf{x}_N, \tilde{y}_{\mathbf{x}_N})\} \equiv \mathcal{D}_\eta$ drawn from $X \times Y_\eta$. 
Assume that label noise is conditionally independent on $X$ (i.e., $\Prob(\tilde{y}_\mathbf{x}|y_\mathbf{x},\mathbf{x}) = \Prob(\tilde{y}_\mathbf{x}|y_\mathbf{x})$), the noisy label is generated as 
\begin{align}
    \tilde{y}_\mathbf{x} = 
    \begin{cases}
        y_\mathbf{x} \quad &\text{with probability } (1-\eta_{y_\mathbf{x}})\\
        j, j \in [K], j \neq y_\mathbf{x} &\text{with probability } \eta_{y_\mathbf{x}j}
    \end{cases} ,
\end{align}
where $\eta_{y_\mathbf{x}j}$ represents the transition probability from the true label $y_\mathbf{x}$ to the noisy label $j$, i.e., $\eta_{y_\mathbf{x} j} = \Prob(Y_\eta = j|Y=y_\mathbf{x})$, and $j \in [K]$ is a concise notation for $j \in \{1, \dots, K\}$. $\eta_{y_\mathbf{x}} = \sum_{j\neq y_\mathbf{x}} \eta_{y_{\mathbf{x}}j}$ is defined as the \emph{noise rate}. 

In Sec. \ref{subsec:obj_fcn_correction} and \ref{subsec:estimator_correction}, we first design an objective function correction approach. Then, we propose a posterior estimator correction method, to achieve label noise robustness. They both rely on the hypothesis of having the transition probabilities $\eta_{y_\mathbf{x} j}$. When the transition probabilities are unknown they can be estimated, as outlined in Sec. \ref{sec:related}.

\subsection{Objective Function Correction}
\label{subsec:obj_fcn_correction}
In this section, we propose an objective function correction approach that leads the training to converge to the same neural network that would be learned using the clean dataset, even in the presence of label noise.
We first study the binary classification case and then extend it to multi-class classification.

\subsubsection{Binary Classification}
Let ${Y}=\{ 0,1\}$ be the labels set. Define the following quantities: $e_0 \triangleq \Prob(Y_\eta=0 | Y=1)$, $e_1 \triangleq \Prob(Y_\eta=1 | Y=0)$ for simplicity in the notation. In the following, we always assume $e_0 + e_1 <1$.
In Theorem \ref{thm:binary_obj_fcn}, we show the effect of label noise on the class of objective functions in \eqref{eq:supervised_general_value_function}.

\begin{theorem}
\label{thm:binary_obj_fcn}
    For the binary classification scenario, the relationship between the value of the objective function in the presence ($\mathcal{J}^\eta_f(T)$) and absence ($\mathcal{J}_f(T)$) of label noise, given the same parametric function $T$, is
    \begin{align}
    \mathcal{J}^\eta_f(T) = (1- e_0 - e_1)\mathcal{J}_f(T) + B_f(T) ,
    \end{align}
    where 
    \begin{align}
    \label{eq:bias_binary}
        B_f(T) \triangleq& \E_{X} \Bigl[ e_0 T(\mathbf{x},0) + e_1 T(\mathbf{x},1) \notag \\
        & - (e_0 + e_1) \sum_{i=0}^1 f^*(T(\mathbf{x}, i)) \Bigr] 
    \end{align}
    is a bias term.
\end{theorem}

In corollary \ref{corollary:binary_obj_fcn_correction}, we show how to perform the objective function correction to remove the effect of label noise.

\begin{corollary}
    \label{corollary:binary_obj_fcn_correction}
    Let us assume the label noise transition probabilities are correctly estimated. Let us define 
    \begin{align}
        \mathcal{J}^{\eta, C}_f(T) \triangleq \mathcal{J}^\eta_f(T) - \hat{B}_f(T) ,
    \end{align}
    where $\hat{B}_f(T)$ is the estimated bias term. Then,
    \begin{align}
        T^\diamond =  \argmax_T \mathcal{J}_f(T) =\argmax_T \mathcal{J}^{\eta, C}_f(T) .
    \end{align}
\end{corollary}
Corollary \ref{corollary:binary_obj_fcn_correction} directly follows from Theorem \ref{thm:binary_obj_fcn}, since if the transition matrix is correctly estimated or known, the bias estimate $\hat{B}_f(T)$ is accurate (i.e., $\hat{B}_f(T) = B_f(T)$). Then, the maximization of $(1- e_0 - e_1)\mathcal{J}_f(T)$ over $T$ is equivalent to the maximization of $\mathcal{J}_f(T)$.

\subsubsection{Multi-class Classification}

Let us first define the notation for the multi-class classification case with asymmetric uniform off-diagonal label noise: $e_j \triangleq P(Y_\eta=j|Y=i) = \eta_{ij} \quad \forall i \neq j$. Assume $\sum_{j \neq i}e_j  < 1$. 

Theorem \ref{thm:multiclass_uniform_obj_fcn} extends Theorem \ref{thm:binary_obj_fcn} for the multi-class case. 
\begin{theorem}
\label{thm:multiclass_uniform_obj_fcn}
    For multi-class asymmetric uniform off-diagonal label noise, the relationship between the value of the objective function in the presence ($\mathcal{J}^\eta_f(T)$) and absence ($\mathcal{J}_f(T)$) of label noise, given the same parametric function $T$, is
    \begin{align}
    \mathcal{J}^\eta_f(T) =& \left( 1 - \sum_{j = 1}^K e_j \right) \mathcal{J}_f(T) + B_f(T) ,
    \end{align}
    where 
    \begin{align}
        B_f(T) \triangleq& \E_X \Biggl[ \sum_{j = 1}^K \Biggl( e_j T(\mathbf{x}, j)  - \left( \sum_{i=1}^K e_i \right) f^*(T(\mathbf{x}, j)) \Biggr) \Biggr] .
    \end{align}
\end{theorem}

Corollary \ref{corollary:binary_obj_fcn_correction} holds true also for the multi-class extension, for the same motivation provided in the binary scenario.


\subsection{Posterior Estimator Correction}
\label{subsec:estimator_correction}
In this section, we present an alternative correction procedure that removes the effect of label noise during the test phase, acting on the posterior estimator obtained by training the neural network with the noisy dataset. 

Let $\hat{p}_{Y|X}$ and $\hat{p}^\eta_{Y|X}$ be the posterior estimators obtained with the clean and noisy datasets, respectively. In general,
\begin{align}
    \hat{y}_\mathbf{x} = \argmax_{y_\mathbf{x}} \hat{p}_{Y|X}(y_\mathbf{x}|\mathbf{x}) \neq \argmax_{y_\mathbf{x}} \hat{p}^\eta_{Y|X}(y_\mathbf{x}|\mathbf{x}) = \hat{y}_{\mathbf{x}}^\eta  .
\end{align}
First, we study the relationship between $\hat{p}_{Y|X}$ and $\hat{p}^\eta_{Y|X}$ by making explicit the effect of label noise in the expression of $\hat{p}^\eta_{Y|X}$, which from \eqref{eq:MAP} depends on 
\begin{align}
    T^{\diamond}_\eta = \argmax_T \mathcal{J}^\eta_f(T) 
\end{align}
instead of $T^\diamond$. Then, we show how to correct the posterior estimate to make it robust to label noise. 


\subsubsection{Binary Classification}

Theorem \ref{thm:binary_posterior} describes the relationship between the posterior estimator in the presence and absence of label noise. 
\begin{theorem}
\label{thm:binary_posterior}
    For the binary classification case, the posterior estimator in the presence of label noise is related to the true posterior as
    \begin{align}
        \hat{p}^\eta_{Y|X}(i|\mathbf{x}) &= (f^*)^\prime(T^{\diamond}_\eta(\mathbf{x}, i)) \notag \\
        &= (1 - e_0 - e_1)p_{Y|X}(i|\mathbf{x}) + e_i ,
    \end{align}
    $\forall i \in \{0,1 \}$.
\end{theorem}

In Corollary \ref{corollary:binary_posterior_correction}, we show how to correct the estimate of the posterior to remove the effect of label noise.

\begin{corollary}
\label{corollary:binary_posterior_correction}
Let us assume the transition probabilities are correctly estimated. Define
\begin{align}
    \hat{p}^{\eta ,C}_{Y|X}(i|\mathbf{x}) \triangleq \hat{p}^\eta_{Y|X}(i|\mathbf{x}) - \hat{e}_i .
\end{align}
Then,
\begin{align}
    \hat{y}_{\mathbf{x}} = \argmax_{y_\mathbf{x} \in \mathcal{A}_y} \hat{p}_{Y|X}(y_\mathbf{x}|\mathbf{x}) = \argmax_{y_{\mathbf{x}} \in \mathcal{A}_y} \hat{p}^{\eta, C}_{Y|X}(y_\mathbf{x}|\mathbf{x}) .
\end{align}
\end{corollary}

Corollary \ref{corollary:binary_posterior_correction} directly follows from Theorem \ref{thm:binary_posterior}.
The estimate of the class is computed by maximizing $\hat{p}^{\eta, C}_{Y|X}(y_\mathbf{x}|\mathbf{x})$ w.r.t. the class element. Therefore, the multiplication by the positive constant does not affect the argmax of the posterior and thus the classification problem can be solved following \eqref{eq:MAP} using $\hat{p}^{\eta, C}_{Y|X}(y_\mathbf{x}|\mathbf{x})$.

\subsubsection{Multi-class Case}
Theorem \ref{thm:multiclass_posterior} extends Theorem \ref{thm:binary_posterior} for the case of asymmetric uniform off-diagonal label noise.
\begin{theorem}
\label{thm:multiclass_posterior}
    For multi-class asymmetric uniform off-diagonal label noise, the relationship between the posterior estimator in the presence of label noise and the true posterior is  
    \begin{align}
    \label{eq:posterior_multiclass}
        \hat{p}^\eta_{Y|X}(i|\mathbf{x}) &= (f^*)^\prime(T^{\diamond}_\eta(\mathbf{x},i)) \notag \\
        &= \left( 1 - \sum_{j = 1}^K e_j \right) p_{Y|X}(i|\mathbf{x}) + e_i ,
    \end{align}
    $\forall i \in \{1,\dots,K \}$.
\end{theorem}


One main difference between the posterior correction and the objective function correction is at what stage of the algorithm they are applied. 
Indeed, from Corollary \ref{corollary:binary_obj_fcn_correction} the bias is removed during training, ensuring that maximizing the objective function is equivalent under both noisy and clean conditions. 
Therefore, the neural network learned in the noisy setting is equal to the one trained in the clean scenario.
Differently, the posterior estimator correction in Corollary \ref{corollary:binary_posterior_correction} is conducted during the test phase. When performing the posterior correction, the neural network learned in noisy conditions differs from the one trained in the clean scenario. However, the bias' subtraction leads to a maximization (w.r.t. the class ${y}_\mathbf{x}$) of the corrected posterior in the noisy setting that is equivalent to the maximization of the posterior in the clean scenario.  

In Section \ref{subsec:robustness_analysis}, we demonstrate the robustness of $f$-PML to symmetric label noise. In such a case, the knowledge of the transition probabilities is not required and $f$-PML is robust without needing any type of correction approach.

\subsection{Robustness Analysis}
\label{subsec:robustness_analysis}

As pointed out in \cite{zhou2023asymmetric}, the majority of robust objective functions are symmetric losses \cite{ghosh2017robust, wang2019symmetric, ma2020normalized, ye2023active}. We demonstrate that, even if the class of objective functions in \eqref{eq:supervised_general_value_function} is not symmetric, it is robust to symmetric label noise, which is a noise model for which researchers showed notable interest \cite{menon2020can}. The noise is defined as \emph{symmetric} if the true label transitions to any other label with equal probability, i.e., $e_j = \frac{\eta}{K-1} = \eta_{ij}, \forall j \neq i$, where $\eta$ is constant. 

A classification algorithm is \textit{noise tolerant} (i.e., robust to label noise) when the classifier learned on noisy data has the same probability of correct classification as the classifier learned on clean data \cite{manwani2013noise}, i.e., 
\begin{align}
\label{eq:noise_tolerance}
    \Prob[pred \circ T^\diamond(\mathbf{x}) = y_\mathbf{x}] = \Prob[pred \circ T^\diamond_\eta(\mathbf{x}) = y_\mathbf{x}] ,
\end{align}
where $pred$ indicates the function used to predict the class. 
In Theorem \ref{thm:robustness}, we prove the robustness of $f$-PML for symmetric label noise under a mild condition on the noise rate. 

\begin{theorem}
    \label{thm:robustness}
    In a multi-class classification task, $f$-PML is noise tolerant under symmetric label noise if $\eta < \frac{K-1}{K}$. 
\end{theorem}

%
Theorem \ref{thm:robustness} guarantees label noise robustness for \emph{any} $f$-divergence and any neural network architecture. 
Usually, the label noise robustness is studied by proving that a certain objective function is symmetric \cite{ghosh2017robust}, which means that the sum of the losses computed over all the classes is constant. 
The objective function symmetry leads to the condition $T^\diamond(\mathbf{x})=T^\diamond_\eta(\mathbf{x})$ \cite{ghosh2017robust, ma2020normalized}, which trivially proves the label noise robustness.
However, that is only a sufficient condition for \eqref{eq:noise_tolerance} to be true. 
Therefore, there can be losses that are robust to label noise but for which $T^\diamond(\mathbf{x}) \neq T^\diamond_\eta(\mathbf{x})$. This is exactly the case of $f$-PML.

\paragraph{Robustness of the CE} Ghosh et al. \cite{ghosh2017robust} showed that the CE is not symmetric. Starting from this statement, some papers analyzed the CE more deeply, studying its gradients \cite{liu2020early}, and highlighting its problem of under learning on some "hard" classes \cite{wang2019symmetric}. On the robustness side, Ghosh et al. could not prove the robustness of the CE, but did not prove its non-robustness, as the symmetry is a sufficient but not necessary condition for the robustness. Misinterpreting Ghosh et al., a series of imprecise statements followed in subsequent papers, that led the CE to be considered as not robust to label noise. For instance, Zhang et al. \cite{zhang2018generalized} write that "Being a nonsymmetric and unbounded loss function, CCE is sensitive to label noise", where CCE stands for categorical cross-entropy. Furthermore, Ma et al. \cite{ma2020normalized} wrongly affirm that the CE is not robust to label noise. Notably, the CE can be obtained from the class of objective functions analyzed in this paper, which is robust to symmetric label noise (see Appendix \ref{sec:appendix_obj_fcns} for more details). Thus, the CE is robust to symmetric label noise.  



\subsection{Convergence Analysis}
\label{subsec:convergence_analysis}
In this section, we study the convergence property of the posterior estimator in the presence of label noise. We provide a theoretical study of the bias between the true posterior (referred to as $p^\diamond$), the posterior estimator attained maximizing $\mathcal{J}^\eta_f(T)$ (referred to as $p^{\diamond}_\eta $), and the estimator obtained in the noisy setting during training (referred to as $p^{{(i)}}_\eta$) without employing any correction approach. 

Theorem \ref{thm:bound_bias} presents a bound on the bias between the posterior estimate at convergence in the presence of label noise and the value of the estimator during training.
\begin{theorem}
\label{thm:bound_bias}
    Let $T^{{(i)}}_\eta$ be the neural network at the $i$-th step of training maximizing $\mathcal{J}^\eta_f(T)$. Assume $T^{{(i)}}_\eta$ belongs to the neighborhood of $T^{\diamond}_\eta$. The bias during training is bounded as
    \begin{align}
    |p^{\diamond}_\eta - p^{{(i)}}_\eta| \leq ||(T^{\diamond}_\eta - T^{{(i)}}_\eta)||_2 ||(f^*)^{\prime \prime}(T^{{(i)}}_\eta)||_2 .
\end{align}
\end{theorem}

Theorem \ref{thm:convergence_multiclass_estimator} describes the bias during training between the optimal posterior estimator and the posterior estimator at the \emph{i}-th iteration of training learned by maximizing the noisy objective function $\mathcal{J}^\eta_f(T)$.
\begin{theorem}
\label{thm:convergence_multiclass_estimator}
    Let $T_{\eta j}^{\diamond}$ and $T_{\eta j}^{(i)}$ be the $j$-th output of the posterior estimator at convergence and at the $i$-th iteration of training, respectively. The difference between the optimal posterior estimate without label noise and the estimate at $i$-th iteration in the presence of label noise reads as
    \begin{align}
        p_j^\diamond - p_{\eta j}^{{(i)}} \simeq \left(\sum_{n=1}^Ke_n\right) p_j^\diamond - e_j + \delta_j^{(i)}(f^*)^{\prime \prime}(T_{\eta j}^{\diamond} - \delta_j^{(i)}),
    \end{align}
    where $\delta_j^{(i)} = T_{\eta j}^{\diamond} - T_{\eta j}^{(i)}$. 
\end{theorem}
Theorems \ref{thm:bound_bias} and \ref{thm:convergence_multiclass_estimator} provide conditions on the biases depending on the $f$-divergence employed, showing that different $f$-divergences lead to diverse biases. 

\subsection{Comparison with Related Work}
\label{subsec:comparison_related}
In \cite{ma2020normalized}, the authors propose the class of active passive losses (APLs), which consists of the sum of an active and a passive loss (see Appendix \ref{subsec:APLs}). 
We observe that $f$-PML resembles the APLs. In fact, the first expectation $\E_{XY}$ is affected only by the neural network's prediction corresponding to the label, while the second expectation $\E_X$ is impacted by the neural network's predictions corresponding to classes different from the label. 
In \cite{ye2023active}, the authors first notice that all the passive losses studied in \cite{ma2020normalized} are based on MAE and then improve the performance of APLs by replacing MAE with different losses. 
In contrast to the explicit APL-based objective function design in \cite{ma2020normalized, ye2023active}, where the active and passive terms are unrelated, the discriminative formulation of the MAP problem of $f$-PML leads to an APL-like objective function which synchronizes the active and passive terms by implicitly considering their interdependency. 
Further details are provided in Appendix \ref{sec:appendix_related}.

A class of objective functions based on the $f$-divergence has also been proposed in \cite{wei2020optimizing}, where the authors maximize the $f$-MI between the classifier's output and label distribution. $f$-PML and the class proposed in \cite{wei2020optimizing} are radically diverse, and we highlight here two main differences. First, $f$-PML returns a Bayes classifier for any $f$, unlike the maximization of the $f$-MI. 
Second, the objective functions proposed in \cite{wei2020optimizing} require sampling from $p_X(\mathbf{x})p_Y(y)$, which is often impractical as we typically only have access to joint samples from $p_{XY}(\mathbf{x},y_\mathbf{x})$. Therefore, the samples from $p_X(\mathbf{x})p_Y(y)$ are often obtained using a shuffling (or derangement) operation which does not guarantee that the resulting samples truly belong to $p_X(\mathbf{x})p_Y(y)$ \cite{mcallester2020formal, letizia2023variational}. Further details are provided in Appendix \ref{sec:appendix_related}. 

\section{Results}
\label{sec:results}

We empirically test $f$-PML for classification in the presence of label noise.
First, we investigate the performance of the proposed correction approaches in binary and multi-class classification scenarios. Then, we evaluate $f$-PML on benchmark datasets for learning with noisy labels.

\paragraph{Baselines} As baselines, we consider the CE, Forward \cite{patrini2017making}, GCE \cite{zhang2018generalized}, SCE \cite{wang2019symmetric}, Co-teaching \cite{han2018co}, Co-teaching+ \cite{yu2019does}, JoCoR \cite{wei2020combating}, ELR \cite{liu2020early}, Peer Loss \cite{liu2020peer}, NCE+RCE/NCE+MAE/NFL+RCE/NFL+MAE \cite{ma2020normalized}, NCE+AEL/NCE+AGCE/NCE+AUL \cite{zhou2021asymmetric}, F-Div \cite{wei2020optimizing}, Divide-Mix \cite{li2020dividemix}, Negative-LS \cite{wei2022smooth},
SOP \cite{liu2022robust}, ProMix \cite{wang2022promix}, ANL-CE/ANL-FL \cite{ye2023active}, RDA \cite{lienen2024mitigating}, SGN \cite{englesson2024robust}. 

\paragraph{Implementation details} Unless differently specified, we use a ResNet34 \cite{resnet} for all the experiments of $f$-PML, consistently with the related work. Optimization is executed using SGD with a momentum of $0.9$. The learning rate is initially set to $0.02$ and a cosine annealing scheduler \cite{loshchilov2016sgdr} decays it during training. 
Since the design of the objective function is orthogonal to the choice of the architecture and training strategy, we additionally test $f$-PML employing the ProMix architecture and training strategy (referring to it as $f$-PML$_{Pro}$), keeping the architecture and hyper-parameters fixed to the values proposed in \cite{wang2022promix}. The tables report the mean over $5$ independent runs of the code with different random seeds. 
Additional details are reported in Appendix \ref{subsec:appendix_implementation_details}. 

\begin{table}[h]
\caption{Test accuracy on breast cancer dataset.} 
  \begin{center}
  \begin{sc}
  \resizebox{\columnwidth}{!}{%
    \begin{tabular}{ c c c c | c } 
     \hline
     Div. & No Cor. & O.F. Cor. & P. Cor. & No Noise \\
     \hline
     KL-PML & $92.10$ & $95.60$ & $95.60$ & $98.20$ \\
     SL-PML & $92.10$ & $95.60$ & $95.60$ & $98.20$ \\
     GAN-PML & $93.00$ & $94.70$ & $95.60$ & $98.20$ \\
     \hline
    \end{tabular}
    }
    \end{sc}
    \end{center}
    \vskip -0.1in
    \label{tab:binary_breast_01_03}
\end{table}

\begin{table}[h]
\caption{Test accuracy on CIFAR-10 for custom transition matrix.} 
  \begin{center}
  \begin{small}
  \begin{sc}
    \begin{tabular}{ c c c c c } 
     \hline
     & \multicolumn{2}{c}{\textbf{Low noise}} & \multicolumn{2}{c}{\textbf{High noise}} \\
     Div. & No Cor. & P. Cor. & No Cor. & P. Cor. \\
     \hline
     KL-PML & $93.04$ & $93.26$ & $83.62$ & $84.66$ \\
     SL-PML & $93.23$ & $92.93$ & $84.80$ & $85.48$ \\
     GAN-PML & $93.03$ & $92.62$ & $84.32$ & $84.90$ \\
     \hline
    \end{tabular}
    \end{sc}
    \end{small}
    \end{center}
    \vskip -0.1in
    \label{tab:cifar10_customT}
\end{table}

\paragraph{Objective function and posterior correction} 
We evaluate the objective function and posterior correction approaches on the breast cancer binary classification dataset \cite{breast_cancer_wisconsin} available on Scikit-learn \cite{pedregosa2011scikit}, and on CIFAR-10 \cite{krizhevsky2009learning} using a custom transition matrix (defined in Appendix \ref{subsubsec:appendix_experiments_correction}). For the binary dataset, the test accuracy achieved using $f$-PML for $e_0=0.1$ and $e_1=0.3$, reported in Tab. \ref{tab:binary_breast_01_03}, shows the performance improvement achieved using the objective function correction (O.F. Cor.) and posterior correction (P. Cor.) approaches. 
We noticed that on average, in practice, the posterior correction approach achieves slightly higher accuracy. 
For CIFAR-10, the comparison between no correction and posterior correction is reported in Tab. \ref{tab:cifar10_customT}. 
Additional results are reported in Appendix \ref{sec:appendix_results}. 

\begin{table}
\caption{Test accuracy of methods with an APL-like objective function, on CIFAR-10 with symmetric noise, using an $8$-layer CNN.} 
  \begin{center}
  \begin{sc}
  \resizebox{\columnwidth}{!}{%
    \begin{tabular}{ c c c c c c } 
    \toprule
    \textbf{Method} & \multicolumn{5}{c}{\textbf{Symmetric}} \\
    \cmidrule{2-6} 
     & Clean & $20\%$ & $40\%$ & $60\%$ & $80\%$ \\
     \toprule
     NFL+MAE & $89.25_{\pm 0.19}$ & $87.33_{\pm 0.14}$ & $83.81_{\pm 0.06}$ & $76.36_{\pm 0.31}$ & $45.23_{\pm 0.52}$ \\
     NFL+RCE & $90.91_{\pm 0.02}$ & $89.14_{\pm 0.13}$ & $86.05_{\pm 0.12}$ & $79.78_{\pm 0.13}$ & $55.06_{\pm 1.08}$ \\ 
     NCE+MAE & $88.83_{\pm 0.34}$ & $87.12_{\pm 0.21}$ & $84.19_{\pm 0.43}$ & $77.61_{\pm 0.05}$ & $49.62_{\pm 0.72}$ \\
     NCE+RCE & $90.76_{\pm 0.22}$ & $89.22_{\pm 0.27}$ & $86.02_{\pm 0.09}$ & $79.78_{\pm 0.50}$ & $52.71_{\pm 1.90}$ \\
     NCE+AEL & $88.51_{\pm 0.26}$ & $86.59_{\pm 0.24}$ & $83.07_{\pm 0.46}$ & $75.06_{\pm 0.26}$ & $41.79_{\pm 1.40}$  \\
     NCE+AGCE & $91.08_{\pm 0.06}$ & $89.11_{\pm 0.07}$ & $86.16_{\pm 0.10}$ & $80.14_{\pm 0.27}$ & $55.62_{\pm 4.78}$ \\
     NCE+AUL & $91.26_{\pm 0.12}$ & $89.08_{\pm 0.14}$ & $86.11_{\pm 0.27}$ & $79.39_{\pm 0.41}$ & $54.49_{\pm 2.77}$  \\
     ANL-CE & $91.66_{\pm 0.04}$ & $90.02_{\pm 0.23}$ & $87.28_{\pm 0.02}$ & $81.12_{\pm 0.30}$ & $61.27_{\pm 0.55}$ \\
     ANL-FL & $91.79_{\pm 0.19}$ & $89.95_{\pm 0.20}$ & $87.25_{\pm 0.11}$ & $81.67_{\pm 0.19}$ & $61.22_{\pm 0.85}$  \\
     \hline
     SL-PML & $\textbf{92.96}_{\pm 0.15}$ & $\textbf{91.16}_{\pm 0.21}$ & $\textbf{87.44}_{\pm 0.19}$ & $81.85_{\pm 0.28}$ & $64.27_{\pm 0.61}$\\
     GAN-PML & $92.92_{\pm 0.09}$ & $90.59_{\pm 0.16}$ & $87.20_{\pm 0.18}$ & $\textbf{82.51}_{\pm 0.23}$ & $\textbf{73.91}_{\pm 0.56}$ \\
     \hline
    \end{tabular}
    }
    \end{sc}
    \end{center}
    \vskip -0.1in
    \label{tab:CIFAR10_symm_APLs}
\end{table}

\begin{table}
\caption{Test accuracy on CIFAR-10 with symmetric noise. All methods use ResNet34.} 
  \begin{center}
  \begin{sc}
  \resizebox{\columnwidth}{!}{%
    \begin{tabular}{ c c c c c } 
    \toprule
    \textbf{Method} & \multicolumn{4}{c}{\textbf{Symmetric}} \\
    \cmidrule{2-5} 
      & $20\%$ & $40\%$ & $60\%$ & $80\%$ \\
     \toprule
     CE & $86.32_{\pm 0.18}$ & $82.65_{\pm 0.16}$ & $76.15_{\pm 0.32}$ & $59.28_{\pm0.97}$ \\
     Forward & $87.99_{\pm0.36}$ & $83.25_{\pm 0.38}$ & $74.96 _{\pm 0.65}$ & $54.64_{\pm 0.44}$  \\
     GCE & $89.83_{\pm 0.20}$ & $87.13_{\pm 0.22}$ & $82.54_{\pm 0.23}$ & $64.07_{\pm 1.38}$ \\
     ELR & $91.16_{\pm0.08}$ & $89.15_{\pm0.17}$ & $86.12_{\pm 0.49}$ & $73.86_{\pm 0.61}$  \\
     SOP & $93.18_{\pm 0.57}$ & $90.09_{\pm 0.27}$ & $\textbf{86.76}_{\pm 0.22}$ & $68.32_{\pm 0.77}$\\
     \hline
     SL-PML &  $92.97_{\pm 0.37}$ & $\textbf{90.38}_{\pm 0.41}$ & $85.25_{\pm 0.44}$ & $65.29_{\pm 0.86}$\\
     GAN-PML & $\textbf{93.20}_{\pm 0.13}$ & $90.05_{\pm 0.21}$ & $84.18_{\pm 0.32}$ & $\textbf{74.91}_{\pm 0.72}$ \\
     \hline
    \end{tabular}
    }
    \end{sc}
    \end{center}
    \vskip -0.1in
    \label{tab:CIFAR10_symm_objective_functions}
\end{table}

\begin{table}[h!]
\caption{Test accuracy on CIFAR-10 and CIFAR-100 with symmetric noise. All methods use refined training strategies.} 
  \begin{center}
  \begin{sc}
  \resizebox{\columnwidth}{!}{%
    \begin{tabular}{ c c c c c c c } 
    \toprule
    \textbf{Method} & \multicolumn{3}{c}{\textbf{CIFAR-10}} & \multicolumn{3}{c}{\textbf{CIFAR-100}} \\
    \cmidrule{2-7} 
     & $20\%$ & $50\%$ & $80\%$ & $20\%$ & $50\%$ & $80\%$ \\
     \toprule
     Co-teaching+ & $89.5$ & $85.7$ & $67.4$ & $65.6$ & $51.8$ & $27.9$\\
     JoCoR & $85.7$ & $79.4$ & $27.8$ & $53.0$ & $43.5$ & $15.5$  \\
     DivideMix & $96.1$ & $94.6$ & $93.2$ & $77.1$ & $74.6$ & $60.2$ \\
     ELR+ & $95.8$ & $94.8$ & $93.3$ & $77.7$ & $73.8$ & $60.8$\\
     SOP+ & $96.3$ & $95.5$ & $94.0$ & $78.8$ & $75.9$ & $63.3$ \\
     ProMix & $97.7$ & $\textbf{97.4}$ & $95.5$ & $\textbf{82.6}$ & $\textbf{80.1}$ & $\textbf{69.4}$ \\
     \hline
     SL-PML$_{Pro}$ & $97.5$ & $96.9$ & $95.6$ & $81.0$ & $77.5$ & $64.4$\\
     GAN-PML$_{Pro}$ & $\textbf{97.8}$ & $97.1$ & $\textbf{96.2}$ & $\textbf{82.6}$ & $79.5$ & $\textbf{69.4}$\\
     \hline
    \end{tabular}
    }
    \end{sc}
    \end{center}
    \vskip -0.1in
    \label{tab:CIFAR10-100_symm_architectures}
\end{table}

\begin{table}[h!]
\caption{Test accuracy on CIFAR-10 and CIFAR-100 with asymmetric noise. An 8-layer CNN is used for CIFAR-10. The ResNet34 is used for CIFAR-100.} 
  \begin{center}
  \begin{sc}
  \resizebox{\columnwidth}{!}{%
    \begin{tabular}{ c c c c c } 
    \toprule
    \textbf{Method} & \multicolumn{2}{c}{\textbf{CIFAR-10}} & \multicolumn{2}{c}{\textbf{CIFAR-100}} \\
    \cmidrule{2-5} 
     & $20\%$ & $30\%$& $20\%$ & $30\%$  \\
     \toprule
     GCE & $85.55_{\pm 0.24}$ & $79.32_{\pm 0.52}$  & $59.06_{\pm 0.46}$ & $53.88_{\pm 0.96}$ \\
     NCE+RCE & $88.36_{\pm 0.13}$ & $84.84_{\pm 0.16}$ & $62.77_{\pm 0.53}$ & $55.62_{\pm 0.56}$  \\
     NCE+AGCE & $88.48_{\pm 0.09}$ & $84.79_{\pm 0.15}$  & $64.05_{\pm 0.25}$ & $56.36_{\pm 0.59}$  \\
     ANL-CE & $89.13_{\pm 0.11}$ & $85.52_{\pm 0.24}$  & $66.27_{\pm 0.19}$ & $59.76_{\pm 0.34}$ \\
     ANL-FL & $89.09_{\pm 0.31}$ & $85.81_{\pm 0.23}$ &  $66.26_{\pm 0.44}$ & $59.68_{\pm 0.86}$  \\
     \hline
     SL-PML & $\textbf{89.14}_{\pm 0.12}$ & $\textbf{86.67}_{\pm 0.27}$  & $70.90_{\pm 0.39}$ & $67.36_{\pm 0.74}$ \\
     GAN-PML & $89.02_{\pm 0.10}$ & $86.14_{\pm 0.21}$ & $\textbf{73.58}_{\pm 0.41}$ & $\textbf{69.80}_{\pm 0.92}$\\
     \hline
    \end{tabular}
    }
    \end{sc}
    \end{center}
    \vskip -0.1in
    \label{tab:CIFAR10-100_asymm}
\end{table}

\begin{table*}[h!]
\caption{Test accuracy achieved on CIFAR-10N and CIFAR-100N.} 
  \begin{center}
  \begin{sc}
  \resizebox{\textwidth}{!}{%
    \begin{tabular}{ c | c c c c c c | c c } 
    \toprule
    \textbf{Method} & \multicolumn{6}{c}{\textbf{CIFAR-10N}} & \multicolumn{2}{c}{\textbf{CIFAR-100N}} \\
     & Clean & Aggregate & Random 1 & Random 2 & Random 3 & Worst & Clean & Noisy\\
     \hline
     CE & $92.92_{\pm 0.11}$ & $87.77_{\pm 0.38}$ & $85.02_{\pm 0.65}$ & $86.46_{\pm 1.79}$ & $85.16_{\pm 0.61}$ & $77.69_{\pm 1.55}$ & $76.70_{\pm 0.74}$ & $55.50_{\pm 0.66}$\\
     Forward & $93.02_{\pm 0.12}$ & $88.24_{\pm 0.22}$ & $86.88_{\pm 0.50}$ & $86.14_{\pm 0.24}$ & $87.04_{\pm 0.35}$ & $79.79_{\pm 0.46}$ & $76.18_{\pm 0.37}$ & $57.01_{\pm 1.03}$\\
     GCE & $92.83_{\pm 0.16}$ & $87.85_{\pm 0.70}$ & $87.61_{\pm 0.28}$ & $87.70_{\pm 0.56}$ & $87.58_{\pm 0.29}$ & $80.66_{\pm 0.35}$ & $76.35_{\pm 0.48}$ & $56.73_{\pm 0.30}$\\
     Co-teaching+ & $92.41_{\pm 0.20}$ & $90.61_{\pm 0.22}$ & $89.70_{\pm 0.27}$ & $89.47_{\pm 0.18}$ & $89.54_{\pm 0.22}$ & $83.26_{\pm 0.17}$ & $70.99_{\pm 0.22}$ & $57.88_{\pm 0.24}$\\
     ELR+ & $95.39_{\pm 0.05}$ & $94.83_{\pm 0.10}$ & $94.43_{\pm 0.41}$ & $94.20_{\pm 0.24}$ & $94.34_{\pm 0.22}$ & $91.09_{\pm 1.60}$ & $78.57_{\pm 0.12}$ & $66.72_{\pm 0.07}$ \\
     Peer Loss & $93.99_{\pm 0.13}$ & $90.75_{\pm 0.25}$ & $89.06_{\pm 0.11}$ & $88.76_{\pm 0.19}$ & $88.57_{\pm 0.09}$ & $82.00_{\pm 0.60}$ & $74.67_{\pm 0.36}$ & $57.59_{\pm 0.61}$ \\
     NCE+RCE & $90.94_{\pm 0.01}$ & $89.17_{\pm 0.28}$ & $87.62_{\pm 0.34}$ & $87.66_{\pm 0.12}$ & $87.70_{\pm 0.18}$ & $79.74_{\pm 0.09}$ & $68.22_{\pm 0.28}$ & $54.27_{\pm 0.09}$ \\
     F-Div & $94.88_{\pm 0.12}$ & $91.64_{\pm 0.34}$ & $89.70_{\pm 0.40}$ & $89.79_{\pm 0.12}$ & $89.55_{\pm 0.49}$ & $82.53_{\pm 0.52}$ & $76.14_{\pm 0.36}$ & $57.10_{\pm 0.65}$ \\
     Divide-Mix & $95.37_{\pm 0.14}$ & $95.01_{\pm 0.71}$ & $95.16_{\pm 0.19}$ & $95.23_{\pm 0.07}$ & $95.21_{\pm 0.14}$ & $92.56_{\pm 0.42}$ & $76.94_{\pm 0.22}$ & $71.13_{\pm 0.48}$ \\
     Negative-LS & $94.92_{\pm 0.25}$ & $91.97_{\pm 0.46}$ & $90.29_{\pm 0.32}$ & $90.37_{\pm 0.12}$ & $90.13_{\pm 0.19}$ & $82.99_{\pm 0.36}$ &$77.06_{\pm 0.73}$ & $58.59_{\pm 0.98}$\\
     JoCoR & $93.40_{\pm 0.24}$ & $91.44_{\pm 0.05}$ & $90.30_{\pm 0.20}$ & $90.21_{\pm 0.19}$ & $90.11_{\pm 0.21}$ & $83.37_{\pm 0.30}$ & $74.07_{\pm 0.33}$ & $59.97_{\pm 0.24}$ \\
     SOP+ & $96.38_{\pm 0.31}$ & $95.61_{\pm 0.13}$ & $95.28_{\pm 0.13}$ & $95.31_{\pm 0.10}$ & $95.39_{\pm 0.11}$ & $93.24_{\pm 0.21}$ & $78.91_{\pm 0.43}$ & $67.81_{\pm 0.23}$ \\
     ProMix & $97.04_{\pm 0.15}$ & $97.65_{\pm 0.19}$ & $97.39_{\pm 0.16}$ & $\textbf{97.55}_{\pm 0.12}$ & $\textbf{97.52}_{\pm 0.09}$ & $96.34_{\pm 0.23}$ & $81.46_{\pm 0.30}$ & $73.79_{\pm 0.28}$ \\
     ANL-CE & $91.66_{\pm 0.04}$ & $89.66_{\pm 0.12}$ & $88.68_{\pm 0.13}$ & $88.19_{\pm 0.08}$ & $88.24_{\pm 0.15}$ & $80.23_{\pm 0.28}$ & $70.68_{\pm 0.23}$ & $56.37_{\pm 0.42}$  \\
     RDA & $94.09_{\pm 0.19}$ & $90.43_{\pm 0.03}$ & $90.09_{\pm 0.29}$ & $90.40_{\pm 0.01}$ & $91.71_{\pm 0.38}$ & $82.91_{\pm 0.83}$ & $76.21_{\pm 0.64}$ & $59.22_{\pm 0.26}$\\
     SGN & $94.12_{\pm 0.22}$ & $92.06_{\pm 0.12}$ & $91.94_{\pm 0.19}$ & $91.69_{\pm 0.22}$ & $91.91_{\pm 0.10}$ & $86.67_{\pm 0.42}$ & $73.88_{\pm 0.34}$ & $60.36_{\pm 0.71}$ \\
     \hline
     SL-PML$_{Pro}$ & $96.08_{\pm 0.20}$ & $97.19_{\pm 0.16}$ & $97.00_{\pm 0.17}$ & $96.93_{\pm 0.09}$ & $97.07_{\pm 0.12}$ & $95.34_{\pm 0.35}$ & $\textbf{82.25}_{\pm 0.45}$ & $72.45_{\pm 0.36}$\\
     GAN-PML$_{Pro}$ & $\textbf{97.20}_{\pm 0.11}$ & $\textbf{97.69}_{\pm 0.21}$ & $\textbf{97.51}_{\pm 0.15}$ & $97.25_{\pm 0.20}$ & $97.30_{\pm 0.13}$ & $\textbf{96.38}_{\pm 0.28}$ & $81.27_{\pm 0.34}$ & $\textbf{73.93}_{\pm 0.29}$\\
     \hline
    \end{tabular}
    }
    \end{sc}
    \end{center}
    \vskip -0.1in
    \label{tab:CIFAR10N_CIFAR100N}
\end{table*}

\begin{table*}[h]
\caption{Test accuracy on ILSVRC12 and Mini WebVision.} 
  \begin{center}
  \begin{sc}
  \resizebox{\textwidth}{!}{%
    \begin{tabular}{ c | c c c c c c c c c } 
     \hline
     Dataset & CE & GCE & SCE & NCE+RCE & NCE+AGCE & ANL-CE & ANL-FL  & GAN-PML & SL-PML \\
     \hline
     ILSVRC12 & $58.64$ & $56.56$ & $62.60$ & $62.40$ & $60.76$ & $65.00$ & $65.56$ & $\textbf{74.56}$ & $74.53$\\
     \hline
     WebVision & $61.20$ & $59.44$ & $68.00$ & $64.92$ & $63.92$ & $67.44$ & $68.32$ & $\textbf{79.53}$ & $77.27$ \\
     \hline
    \end{tabular}
    }
    \end{sc}
    \end{center}
    \vskip -0.1in
    \label{tab:large_real_datasets}
\end{table*}

\paragraph{Comparison with APL-like losses} 
Since $f$-PML possesses APL-like properties, we perform a comparative analysis with the existing APLs and ANLs. 
The comparison in Tab. \ref{tab:CIFAR10_symm_APLs} highlights $f$-PML's superior accuracy (up to $12\%$) over other APL-like methods on symmetric label noise. As the label noise is symmetric, $f$-PML does not need any correction technique. The efficacy of $f$-PML with respect to the other APL-like methods can be attributed to the fact that other APL-like methods rely on the sum of two independent active and passive losses, while the $f$-PML framework implicitly defines a relationship between active and passive terms. In particular, for CIFAR-10, $f$-PML consistently outperforms the other methods. 

\paragraph{Synthetic and realistic label noise} 

We evaluate $f$-PML for the case of synthetic and realistic label noise. 
For symmetric label noise, Tab. \ref{tab:CIFAR10_symm_objective_functions} shows that $f$-PML is also competitive with well-known algorithms for classification with label noise that do not use APL-like losses. Tab. \ref{tab:CIFAR10-100_symm_architectures} compares algorithms using complex architectures and convoluted training strategies with $f$-PML$_{Pro}$, showing that $f$-PML can be used to replace the CE (or other objective functions) to train state-of-the-art architectures. 

For asymmetric label noise, the test accuracy is reported in Tab. \ref{tab:CIFAR10-100_asymm}. Since the asymmetric label noise used for CIFAR-10 and CIFAR-100 is not uniform off-diagonal, we utilize $f$-PML without correction. As for Tabs. \ref{tab:CIFAR10_symm_APLs} and \ref{tab:CIFAR10_symm_objective_functions}, $f$-PML performs better than existing APL-like objective functions and other different techniques.

For the case of realistic label noise, the test accuracy is reported in Tab. \ref{tab:CIFAR10N_CIFAR100N}. Also for the case of realistic label noise, $f$-PML is used without correction techniques, as the label noise model is unknown. 
Even if for $f$-PML$_{Pro}$ we use the same hyperparameters used in ProMix, which are optimal for the CE and have not been refined for other $f$-divergences, in many scenarios $f$-PML$_{Pro}$ achieves the highest performance. 
The numerical results demonstrate the effectiveness of $f$-PML$_{Pro}$, showing that by merging $f$-PML and complex architectures and training strategies it is possible to attain a performance comparable or superior to the state-of-the-art. 

We train a ResNet50 on mini WebVision \cite{li2017webvision} and then test the trained network on the validation datasets of mini WebVision and ImageNet ILSVRC12 \cite{krizhevsky2012imagenet} (Tab. \ref{tab:large_real_datasets}). The training lasts for $100$ epochs and we use a batch size of $64$, with SGD with momentum $0.9$ cosine annealing scheduler and initial learning rate $0.02$. Even if we train $f$-PML on a smaller number of epochs with respect to the other algorithms in Tab. \ref{tab:large_real_datasets}, $f$-PML achieves a significantly higher accuracy.

Additional results are reported in Appendix \ref{sec:appendix_results}.


\section{Conclusions}
\label{sec:conclusions}
In this paper, we analyze an $f$-divergence based posterior maximization learning ($f$-PML) technique for classification with label noise. We propose an objective function correction approach and a novel posterior estimator correction technique to make $f$-PML robust to label noise. Furthermore, we show that $f$-PML is robust to symmetric label noise for any $f$-divergence, without requiring the knowledge of the noise rates. Finally, the experimental results demonstrate the effectiveness of $f$-PML in its simplest form or when it is used in combination with refined training strategies.


\bibliography{main_a}
\bibliographystyle{unsrt}

\newpage
\onecolumn
\appendix
\section{Additional Details on the Objective Functions}
\label{sec:appendix_obj_fcns}


\begin{table}
\caption{$f$-divergences table. The corresponding $f$-divergences are: Kullback-Leibler, 
GAN, 
Shifted Log.} 
\centering
\vskip 0.15in
  \begin{center}
  \begin{small}
  \begin{sc}
    \begin{tabular}{ c c c c } 
     \hline
     Name & $f(u)$ & $f^{*}(t)$ & $T^\diamond(p_{Y|X})$ \\
     \hline
       KL & $u \log(u)$ & $\exp{(t-1)}$ & $\log(p_{Y|X}) + 1$ \\
      GAN & $u \log(u) - (u+1)\log(u+1)$ & $-\log(1-\exp{(t)})$ & $\log(p_{Y|X}/(p_{Y|X}+1))$ \\
      SL & $-\log(u+1)$ & $-(\log(-t) + t)$ & $ -1/(p_{Y|X} +1)$ \\
    \end{tabular}
    \end{sc}
    \end{small}
    \end{center}
    \vskip -0.1in
    \label{tab:f_divergences_table}
\end{table}

The generator functions of the $f$-divergences used in this paper are reported in Tab. \ref{tab:f_divergences_table}, along with their Fenchel conjugate functions $f^*$, and the optimal value achieved by the neural network at convergence $T^\diamond = f^\prime(p_{XY}/p_X)$.

In the following, we list the objective functions of $f$-PML corresponding to different $f$-divergences.
For the experiments on the objective function and posterior correction approaches, following the work in \cite{f_gan}, the neural network's output is expressed as $T = g_f(v)$, where $v$ is a linear layer output of the neural network, and $g_f(\cdot)$ is a monotonically increasing function as defined in \cite{f_gan}. However, we noticed that for datasets with a large amount of classes, like CIFAR-100, the training sometimes fails when using these objective functions. 
Therefore, for those datasets, we apply a change of variable $T=r(D)$ that improves the training process, where $T$ is not expressed based on $g_f(\cdot)$. For all the objective functions, we use the following notation: $\textbf{1}_K(y_\mathbf{x})$ is a one-hot column vector equal to $1$ in correspondence of the label $y_\mathbf{x}$, $\textbf{1}_K$ is a column vector of $1$s of length $K$.

\paragraph{Kullback-Leibler divergence} The objective function corresponding to the KL divergence is
\begin{align}
\label{eq:KL_T}
    \mathcal{J}_{KL}(T) = \E_{{XY}}\left[ T(\mathbf{x})\textbf{1}_K(y_\mathbf{x}) \right] + \E_{X}\left[ \sum_{i=1}^K - e^{T(\mathbf{x}, i) - 1} \right] .
\end{align}

Substituting $T^\diamond$ from Tab. \ref{tab:f_divergences_table}, we get
\begin{align}
    \mathcal{J}_{KL}(T^\diamond) = \E_{{XY}}\left[ \log\left( p_{Y|X}(y_\mathbf{x}|\mathbf{x}) \right) \right] + \E_{X}\left[ \sum_{i=1}^K \left(  - p_{Y|X}(i|\mathbf{x}) \right) \right] .
\end{align}

\textbf{Robustness to symmetric label noise of the CE}
Using the change of variable $T=\log(D)+1$ (thus $D(\mathbf{x})=[p_{Y|X}(1|\mathbf{x}), \dots, p_{Y|X}(K|\mathbf{x})]$), the objective function rewrites as
\begin{align}
    \mathcal{J}_{KL}(D) = \E_{XY}\left[ \log(D(\mathbf{x}))\textbf{1}_K(y_\mathbf{x}) \right] + \E_X\left[ -D(\mathbf{x})\textbf{1}_K \right].
\end{align}
When using the softmax activation function as output of the neural network, $\E_X\left[ -D(\mathbf{x})\textbf{1}_K \right] = -1$, as $D(\mathbf{x})\textbf{1}_K = \sum_{i=1}^K D(\mathbf{x}, i) = 1$. Thus, $\mathcal{J}_{KL}(D) = \E_{XY}[\log(D(\mathbf{x}))\mathbf{1}_K(y_\mathbf{x})]$, whose maximization is exactly the minimization of the CE. Since the CE belongs to the class $f$-PML, it is robust to symmetric label noise for Theorem \ref{thm:robustness}. 

\paragraph{GAN divergence} The objective function corresponding to the GAN divergence is 
\begin{align}
    \mathcal{J}_{GAN}(T) = \E_{{XY}}\left[ T(\mathbf{x})\textbf{1}_K(y_\mathbf{x}) \right] + \E_{X}\left[ \sum_{i=1}^K \log \left( 1- e^{T(\mathbf{x}, i)}\right) \right] ,
\end{align}
Substituting $T^\diamond$ from Tab. \ref{tab:f_divergences_table}, we get
\begin{align}
    \mathcal{J}_{GAN}(T^\diamond) = \E_{{XY}}\left[ \log\left( \frac{p_{Y|X}(y_\mathbf{x}|\mathbf{x})}{p_{Y|X}(y_\mathbf{x}|\mathbf{x}) + 1} \right) \right] + \E_{X}\left[ \sum_{i=1}^K \log \left( \frac{1}{p_{Y|X}(i|\mathbf{x}) + 1} \right) \right] .
\end{align}
Using the change of variable $T = \log(D/(D+1))$, the objective function writes as
\begin{align}
    \mathcal{J}_{GAN}(D) = \E_{{XY}}\left[ \log\left( \frac{D(\mathbf{x})}{D(\mathbf{x}) + 1} \right) \textbf{1}_K(y_\mathbf{x})\right] + \E_{X}\left[ \sum_{i=1}^K \log \left( \frac{1}{D(\mathbf{x},i) + 1} \right) \right] .
\end{align}

\paragraph{Shifted log divergence} The objective function corresponding to the SL divergence is 
\begin{align}
    \mathcal{J}_{SL}(T) = \E_{{XY}}\left[ T(\mathbf{x})\textbf{1}_K(y_\mathbf{x}) \right] + \E_{X}\left[ -\sum_{i=1}^K \left( -(\log(-T(\mathbf{x}, i) ) + T(\mathbf{x}, i)) \right)  \right] .
\end{align}
Substituting $T^\diamond$ from Tab. \ref{tab:f_divergences_table}, we get
\begin{align}
    \mathcal{J}_{SL}(T^\diamond) = \E_{{XY}}\left[ - \frac{1}{p_{Y|X}(y_\mathbf{x}|\mathbf{x}) + 1} \right] + \E_{X}\left[ \sum_{i=1}^K \left( - \frac{1}{p_{Y|X}(i|\mathbf{x}) + 1} + \log \left( \frac{1}{p_{Y|X}(i|\mathbf{x}) + 1} \right) \right) \right] .
\end{align}
Using the change of variable $T=-1/(D+1)$, the objective function writes as
\begin{align}
    \mathcal{J}_{SL}(D) = \E_{{XY}}\left[ - \frac{1}{D(\mathbf{x}) + 1} \textbf{1}_K(y_\mathbf{x}) \right] + \E_{X}\left[ \sum_{i=1}^K \left( - \frac{1}{D(\mathbf{x}, i) + 1} + \log \left( \frac{1}{D(\mathbf{x},i) + 1} \right) \right) \right] .
\end{align}

\section{Proofs}
\label{sec:appendix_proofs}

\subsection{Proof of Theorem \ref{thm:binary_obj_fcn}}

\begin{appthm}{\ref{thm:binary_obj_fcn}}
    For the binary classification scenario, the relationship between the value of the objective function in the presence ($\mathcal{J}^\eta_f(T)$) and absence ($\mathcal{J}_f(T)$) of label noise, given the same parametric function $T$, is
    \begin{align}
    \mathcal{J}^\eta_f(T) = (1- e_0 - e_1)\mathcal{J}_f(T) + B_f(T) ,
    \end{align}
    where 
    \begin{align}
        B_f(T) \triangleq& \E_{X} \Bigl[ e_0 T(\mathbf{x},0) + e_1 T(\mathbf{x},1) - (e_0 + e_1) \sum_{i=0}^1 f^*(T(\mathbf{x}, i)) \Bigr] 
    \end{align}
    is a bias term.
\end{appthm}
\begin{proof}
The value of the objective function in the presence of label noise, according to \eqref{eq:supervised_general_value_function}, is obtained as 
\begin{align}
    \label{eq:supervised_noisy_general_value_function}
    \mathcal{J}^\eta_f(T) = \E_{{XY_\eta}}\Bigl[ T(\mathbf{x},\tilde{y}_\mathbf{x}) \Bigr] - \E_{X}\Bigl[ \sum_{i=0}^1 f^*(T(\mathbf{x},i)) \Bigr].
\end{align}
Given that the label noise is conditionally independent on $X$, the first term in \eqref{eq:supervised_noisy_general_value_function} rewrites as 
    \begin{align}
    \E_{{XY_\eta}}\left[ T(\mathbf{x}, \tilde{{y}}_\mathbf{x}) \right] =& \E_{Y}\E_{{X|Y}}\E_{{Y_\eta|Y}}\Bigl[ T(\mathbf{x},\tilde{y}_\mathbf{x}) \Bigr]\\ 
    =& p_Y(0) \E_{{X|Y=0}} \left[ \Prob[Y_\eta=0|Y=0] T(\mathbf{x},0) + \Prob[Y_\eta=1|Y=0] T(\mathbf{x},1) \right] \notag \\
    &+ (1-p_Y(0))\E_{{X|Y=1}} \left[ \Prob[Y_\eta=0|Y=1] T(\mathbf{x},0) + \Prob[Y_\eta=1|Y=1] T(\mathbf{x},1) \right] \\
    =& p_Y(0) \E_{{X|Y=0}} \left[ (1-e_1) T(\mathbf{x},0) + e_1 T(\mathbf{x},1) \right] \notag \\
    &+ (1-p_Y(0))\E_{{X|Y=1}} \left[ e_0 T(\mathbf{x},0) + (1-e_0) T(\mathbf{x},1) \right] \\
    =& p_Y(0) \E_{{X|Y=0}} \left[ (1-e_0-e_1) T(\mathbf{x},0) + e_0 T(\mathbf{x},0) + e_1 T(\mathbf{x},1) \right] \notag \\
    &+ (1-p_Y(0))\E_{{X|Y=1}} \left[ e_0 T(\mathbf{x},0) + (1-e_1-e_0) T(\mathbf{x},1) + e_1 T(\mathbf{x},1) \right] \\
    =& p_Y(0) \E_{{X|Y=0}} \left[ (1-e_0-e_1) T(\mathbf{x},0) \right] + (1-p_Y(0)) \E_{{X|Y=1}} \left[ (1-e_0-e_1) T(\mathbf{x},1) \right] \notag \\
    &+ p_Y(0) \E_{{X|Y=0}} \left[ e_0 T(\mathbf{x},0) + e_1 T(\mathbf{x},1)  \right] + (1-p_Y(0)) \E_{{X|Y=1}} \left[ e_0 T(\mathbf{x},0) + e_1 T(\mathbf{x},1) \right] \\
    =& (1-e_0-e_1) \E_{{XY}} \left[ T(\mathbf{x}, {y}_\mathbf{x}) \right] + \E_{{X}} \left[ e_0 T(\mathbf{x},0) + e_1 T(\mathbf{x},1) \right] \label{eq:binary_loss_noisy_obj_fcn_first_term}
\end{align}

and 

\begin{align}
    \E_{{X}}\left[f^*(T(\mathbf{x}, 0)) + f^*(T(\mathbf{x}, 1))\right] =& (1-e_0-e_1) \E_{{X}}\left[ f^*(T(\mathbf{x}, 0)) + f^*(T(\mathbf{x}, 1)) \right] \notag \\
    &+ (e_0 + e_1) \E_{{X}}\left[ f^*(T(\mathbf{x}, 0)) + f^*(T(\mathbf{x}, 1)) \right] .\label{eq:binary_loss_noisy_obj_fcn_sec_term}
\end{align}
The second term is not affected by the presence of label noise.


Subtracting the first RHS term in \eqref{eq:binary_loss_noisy_obj_fcn_sec_term} to the first RHS term in \eqref{eq:binary_loss_noisy_obj_fcn_first_term}, we get
\begin{align}
    (1-e_0-e_1) \E_{{XY}} \left[ T(\mathbf{x},y_\mathbf{x}) \right] - (1-e_0-e_1) \E_{{X}}\left[ \sum_{i=0}^1 f^*(T(\mathbf{x}, i)) \right] = (1-e_0 -e_1) \mathcal{J}_f(T) ,
\end{align}
where $\mathcal{J}_f(T)$ is the value of the objective function when the training is done in the absence of label noise. 
Subtracting the second RHS term in \eqref{eq:binary_loss_noisy_obj_fcn_sec_term} to the second RHS term in \eqref{eq:binary_loss_noisy_obj_fcn_first_term}, we get
\begin{align}
    \E_{{X}} \left[ e_0 T(\mathbf{x},0) + e_1 T(\mathbf{x},1) - (e_0 + e_1) \sum_{i=0}^1 f^*(T(\mathbf{x}, i)) \right] \triangleq B_f(T)
\end{align}
Putting all together, we obtain the theorem's claim.

\end{proof}

\subsection{Proof of Theorem \ref{thm:multiclass_uniform_obj_fcn}}

\begin{appthm}{\ref{thm:multiclass_uniform_obj_fcn}}
    For multi-class asymmetric uniform off-diagonal label noise, the relationship between the value of the objective function in the presence ($\mathcal{J}^\eta_f(T)$) and absence ($\mathcal{J}_f(T)$) of label noise, given the same parametric function $T$, is
    \begin{align}
    \mathcal{J}^\eta_f(T) =& \left( 1 - \sum_{j = 1}^K e_j \right) \mathcal{J}_f(T) + B_f(T) ,
    \end{align}
    where 
    \begin{align}
        B_f(T) \triangleq& \E_X \Biggl[ \sum_{j = 1}^K \Biggl( e_j T(\mathbf{x}, j)  - \Biggl(\sum_{i=1}^K e_i\Biggr) f^*(T(\mathbf{x}, j)) \Biggr) \Biggr] .
    \end{align}
\end{appthm}
\begin{proof}
    Let $p_i \triangleq P(Y=i)$. 
    We have $\tilde{p}_i \triangleq P(\tilde{Y} = i) = \left( 1 - \sum_{j \neq i}e_j \right) p_i + e_i \sum_{j \neq i}p_j$. 
    The objective function in the presence of label noise is
    \begin{align}
    \label{eq:appendix_obj_fcn_multiclass_noisy}
        \mathcal{J}^\eta_f(T) = \E_{{XY_\eta}}\Bigl[ T(\mathbf{x},\tilde{y}_\mathbf{x}) \Bigr] - \E_{X}\Biggl[ \sum_{i=1}^K f^*(T(\mathbf{x},i)) \Biggr].
    \end{align}
    The first term can be rewritten as
\begin{align}
    \E_{{XY_\eta}} \left[ T(\mathbf{x},\tilde{y}_\mathbf{x}) \right] &= \E_Y \E_{X|Y} \E_{{Y_\eta|Y}} \left[ T(\mathbf{x}, \tilde{y}_\mathbf{x}) \right] \\
    &= \sum_{i=1}^K p_i \E_{X|Y=i}\left[ \left( 1 - \sum_{j \neq i} e_j \right)T(\mathbf{x}, i) + \sum_{j \neq i} e_j T(\mathbf{x}, j) \right] \\
    &= \sum_{i=1}^K p_i \E_{X|Y=i}\left[ \left( 1 - \sum_{j = 1}^K e_j \right)T(\mathbf{x}, i) + \sum_{j = 1}^K e_j T(\mathbf{x}, j) \right] \\
    &= \left( 1 - \sum_{j = 1}^K e_j \right)\E_{XY}\left[ T(\mathbf{x}, y_\mathbf{x}) \right] + \sum_{j = 1}^K e_j \E_X \left[T(\mathbf{x}, j) \right].
\end{align}

As in the binary case, the second term of \eqref{eq:appendix_obj_fcn_multiclass_noisy} is not influenced by the presence of label noise.
Merging the two terms we obtain the theorem's claim
\begin{align}
    \mathcal{J}^\eta_f(T) =& \left( 1 - \sum_{j = 1}^K e_j \right)\E_{XY}\left[ T(\mathbf{x}, y_\mathbf{x}) \right] + \sum_{j = 1}^K e_j \E_X \left[T(\mathbf{x}, j) \right] - \E_{X}\Bigl[ \sum_{j=1}^K f^*(T(\mathbf{x}, j)) \Bigr] \\
    =& \left( 1 - \sum_{j = 1}^K e_j \right)\E_{XY}\left[ T(\mathbf{x}, y_\mathbf{x}) \right] - \Bigl( 1 - \sum_{j=1}^K e_j \Bigr)\E_{X}\Bigl[ \sum_{j=1}^K f^*(T(\mathbf{x}, j)) \Bigr] \notag \\
    &+ \underbrace{\sum_{j = 1}^K (e_j \E_X \left[T(\mathbf{x}, j) \right]) - \left(\sum_{j=1}^K e_j \right) \E_{X}\Biggl[ \sum_{j=1}^K f^*(T(\mathbf{x}, j)) \Biggr]}_{\triangleq B_f(T)} \\
    =& \left( 1 - \sum_{j = 1}^K e_j \right) \mathcal{J}_f(T) + B_f(T).
\end{align}

\end{proof}

\subsection{Proof of Theorem \ref{thm:binary_posterior}}

\begin{appthm}{\ref{thm:binary_posterior}}
    For the binary classification case, the posterior estimator in the presence of label noise is related to the true posterior as
    \begin{align}
        \hat{p}^\eta_{Y|X}(i|\mathbf{x}) &= (f^*)^\prime(T_\eta^{\diamond}(\mathbf{x}, i)) \notag \\
        &= (1 - e_0 - e_1)p_{Y|X}(i|\mathbf{x}) + e_i ,
    \end{align}
    $\forall i \in \{0,1 \}$.
\end{appthm}
\begin{proof}
The expression of $\mathcal{J}_f(T)$ can be rewritten as
\begin{align}
    \mathcal{J}_f(T) &= \E_{{XY}}\left[ T(\mathbf{x},y_\mathbf{x}) \right] - \E_{X}\left[ \sum_{i=1}^2 f^*(T(\mathbf{x},i)) \right] \\
    &= \E_{{XY}}\left[ T(\mathbf{x},y_\mathbf{x}) \right] - \E_{X}\left[ f^*(T(\mathbf{x},0)) + f^*(T(\mathbf{x},1)) \right] \\
    &= \E_{Y} \left[ \E_{{X|Y}}\left[ T(\mathbf{x},y_\mathbf{x}) \right] \right] - \E_{X}\left[ f^*(T(\mathbf{x},0)) + f^*(T(\mathbf{x},1)) \right] \\
    &= p_Y(0) \left[ \E_{{X|Y=0}}\left[ T(\mathbf{x},0) \right] \right] + p_Y(1) \left[ \E_{{X|Y=1}}\left[ T(\mathbf{x},1) \right] \right] - \E_{X}\left[ f^*(T(\mathbf{x},0)) + f^*(T(\mathbf{x},1)) \right] \\
    &= \underbrace{p_Y(0) \left[ \E_{{X|Y=0}}\left[ T(\mathbf{x},0) \right] \right] - \E_{X}\left[ f^*(T(\mathbf{x},0)) \right]}_{\triangleq \mathcal{J}_{f,0}(T)} + \underbrace{p_Y(1) \left[ \E_{{X|Y=1}}\left[ T(\mathbf{x},1) \right] \right] - \E_{X}\left[ f^*(T(\mathbf{x},1)) \right] }_{ \triangleq \mathcal{J}_{f,1}(T)}.
\end{align}

Similarly, the bias term in \eqref{eq:bias_binary} can be rewritten as
\begin{align}
    B_f(T) &= \E_{X}\left[ e_0 T(\mathbf{x},0) + e_1 T(\mathbf{x},1) - (e_0 + e_1) \bigl(f^*(T(\mathbf{x},0)) + f^*(T(\mathbf{x},1))\bigr) \right] \\
    &= \underbrace{\E_{X}\left[ e_0 T(\mathbf{x},0) - (e_0 + e_1) f^*(T(\mathbf{x},0)) \right]}_{\triangleq B_{f,0}(T)} + \underbrace{\E_{X}\left[ e_1 T(\mathbf{x},1) - (e_0 + e_1) f^*(T(\mathbf{x},1)) \right]}_{\triangleq B_{f,1}(T)}.
\end{align}

Merging the two expressions for $\mathcal{J}_f$ and $B_f$ with Theorem \ref{thm:binary_obj_fcn}, the objective function in presence of label noise becomes
\begin{align}
    \mathcal{J}^\eta_f(T) &= (1- e_0 - e_1)\mathcal{J}_f(T) + B_f(T) \\
    &= (1- e_0 - e_1) (\mathcal{J}_{f,0}(T) + \mathcal{J}_{f,1}(T)) + B_{f,0}(T) + B_{f,1}(T) \\
    &= \underbrace{(1 - e_0 - e_1)\mathcal{J}_{f,0}(T) + B_{f,0}(T)}_{\triangleq \mathcal{J}^\eta_{f,0}(T)} + \underbrace{(1 - e_0 - e_1)\mathcal{J}_{f,1}(T) + B_{f,1}(T)}_{\triangleq \mathcal{J}^\eta_{f,1}(T)}.
\end{align}

$B_{f,0}(T)$ and $(1 - e_0 - e_1)\mathcal{J}_{f,0}(T)$ are concave in $T$. 
Therefore, $\mathcal{J}^\eta_{f,0}(T)$ is concave in $T$ because sum of concave functions.
Since $\mathcal{J}^\eta_{f,0}(T)$ is concave, the optimal convergence condition of $T$ is achieved imposing the first derivative of $\mathcal{J}^\eta_{f,0}(T)$ equal to $0$. $\mathcal{J}^\eta_{f,0}(T)$ can be rewritten as

\begin{align}
    \mathcal{J}^\eta_{f,0}(T) &= (1 - e_0 - e_1)(p_Y(0) \left[ \E_{{X|Y=0}}\left[ T(\mathbf{x},0) \right] \right] - \E_{X}\left[ f^*(T(\mathbf{x},0)) \right]) + \E_{X}\left[ e_0 T(\mathbf{x},0) - (e_0 + e_1) f^*(T(\mathbf{x},0)) \right] \\
    &= \int_{\mathcal{X}}(1 - e_0 - e_1) (p_Y(0) p_{X|Y}(\mathbf{x}|0) T(\mathbf{x},0) - p_X(\mathbf{x}) f^*(T(\mathbf{x},0))) \notag \\
    & \quad \quad \quad + p_X(\mathbf{x}) e_0 T(\mathbf{x},0) - p_X(\mathbf{x})(e_0 + e_1)f^*(T(\mathbf{x},0)) d\mathbf{x} .
\end{align}

Thus, imposing the first derivative w.r.t. $T$ equals to $0$ yields

\begin{align}
    & (f^*)^\prime(T(\mathbf{x},0)) = (1 - e_0 - e_1)p_{Y|X}(0|\mathbf{x}) + e_0  .
\end{align}

Since $(f^*)^\prime(t) = (f^\prime)^{-1}(t)$,
\begin{align}
    T^\diamond_\eta(\mathbf{x},0) = f^\prime((1 - e_0 - e_1)p_{Y|X}(0|\mathbf{x}) + e_0),
\end{align}
where $T^\diamond_\eta(\mathbf{x},0)$ indicates the neural network at convergence.
Therefore, the posterior estimator obtained in the presence of label noise reads as
\begin{align}
    \hat{p}^\eta_{Y|X}(0|\mathbf{x}) = (f^*)^\prime(T^\diamond_\eta(\mathbf{x},0)) = (1 - e_0 - e_1)p_{Y|X}(0|\mathbf{x}) + e_0 .
\end{align}




The same calculations can be done for $\mathcal{J}^\eta_{f,1}(T)$, leading to
\begin{align}
    \hat{p}^\eta_{Y|X}(1|\mathbf{x}) = (f^*)^\prime(T^\diamond_\eta(\mathbf{x},1)) = (1 - e_0 - e_1)p_{Y|X}(1|\mathbf{x}) + e_1 .
\end{align}

\end{proof}

\subsection{Proof of Theorem \ref{thm:multiclass_posterior}}

\begin{appthm}{\ref{thm:multiclass_posterior}}
    For multi-class asymmetric uniform off-diagonal label noise, the relationship between the posterior estimator in the presence of label noise and the true posterior is  
    \begin{align}
        \hat{p}^\eta_{Y|X}(i|\mathbf{x}) &= (f^*)^\prime(T_\eta^{\diamond}(\mathbf{x},i)) \notag \\
        &= \left( 1 - \sum_{j = 1}^K e_j \right) p_{Y|X}(i|\mathbf{x}) + e_i ,
    \end{align}
    $\forall i \in \{1,\dots,K \}$.
\end{appthm}
\begin{proof}
Similarly to the proof of Theorem \ref{thm:binary_posterior}, $\mathcal{J}_f(T)$ rewrites as 
\begin{align}
    \mathcal{J}_f(T) &= \E_{XY}\Bigl[ T(\mathbf{x},y_\mathbf{x}) \Bigr] - \E_{X}\Bigl[ \sum_{j=1}^K f^*(T(\mathbf{x},j)) \Bigr] \\
    &= \sum_{j=1}^K \Bigl( p_Y(j) \E_{X|Y}\Bigl[ T(\mathbf{x}, j) \Bigr] - \E_{X}\Bigl[ f^*(T(\mathbf{x},j)) \Bigr] \Bigr) \\
    &= \sum_{j=1}^K \mathcal{J}_{f,j}(T)
\end{align}
Analogously, for the bias we obtain
\begin{align}
    B_f(T) &= \sum_{j = 1}^K (e_j \E_X \left[T(\mathbf{x}, j) \right]) - \left(\sum_{i=1}^K e_i\right) \E_{X}\Bigl[ \sum_{j=1}^K f^*(T(\mathbf{x}, j)) \Bigr] \\
    &= \sum_{j=1}^K \Bigl( \E_X\Bigl[ e_jT(\mathbf{x}, j) - \left(\sum_{i=1}^K e_i\right) f^*(T(\mathbf{x}, j)) \Bigr] \Bigr) \\
    &= \sum_{j=1}^K B_{f,j}(T) .
\end{align}

Putting everything together, we obtain
\begin{align}
    \mathcal{J}^\eta_f(T) &= \left( 1 - \sum_{i = 1}^K e_i \right) \mathcal{J}_f(T) + B_f(T) \\
    &= \left( 1 - \sum_{i = 1}^K e_i \right) \sum_{j=1}^K\mathcal{J}_{f,j}(T) + \sum_{j=1}^KB_{f,j}(T) \\
    &= \sum_{j=1}^K \underbrace{\left( \left( 1 - \sum_{i = 1}^K e_i \right) \mathcal{J}_{f,j}(T) + B_{f,j}(T) \right)}_{\triangleq \mathcal{J}^\eta_{f,j}(T)} \\
    &= \sum_{j=1}^K \mathcal{J}^\eta_{f,j}(T)
\end{align}
For the same motivation explained for the binary case, $\mathcal{J}^\eta_{f,j}(T)$ is a concave function of $T$.
Therefore, the optimal convergence of $T$ is achieved imposing the first derivative of $\mathcal{J}_{f,j}^\eta(T)$ equal to zero 
\begin{align}
    &\frac{\partial}{\partial T}\mathcal{J}^\eta_{f,j}(T) = 0 \Rightarrow \\
    &\frac{\partial}{\partial T} \Biggl( \int_{\mathcal{T}_x} \left( 1 - \sum_{i = 1}^K e_i \right) \left( p_Y(j) p_{X|Y}(\mathbf{x}|j) T(\mathbf{x}, j) - p_X(\mathbf{x}) f^*(T(\mathbf{x},j)) \right) + \\
    & \quad \quad \quad \quad \quad + p_X(\mathbf{x}) e_jT(\mathbf{x}, j) - p_X(\mathbf{x})\left(\sum_{i=1}^K e_i\right) f^*(T(\mathbf{x}, j)) d\mathbf{x} \Biggr) = 0 
\end{align}
which implies 
\begin{align}
    & \left( 1 - \sum_{i = 1}^K e_i \right) \left( p_Y(j) p_{X|Y}(\mathbf{x}|j)  - p_X(\mathbf{x}) (f^*)^\prime(T(\mathbf{x},j)) \right) + p_X(\mathbf{x}) e_j - p_X(\mathbf{x}) \left( \sum_{i=1}^K e_i \right) (f^*)^\prime(T(\mathbf{x}, j)) = 0 \\
    &\Rightarrow \left( 1 - \sum_{i = 1}^K e_i \right) p_{XY}(\mathbf{x},j) + p_X(\mathbf{x})e_j = p_X(\mathbf{x})(f^*)^\prime(T(\mathbf{x},j)) \\
    & \Rightarrow \left( 1 - \sum_{i = 1}^K e_i \right) p_{Y|X}(j|\mathbf{x}) + e_j = (f^*)^\prime(T(\mathbf{x},j)) .
\end{align}
Since $(f^*)^\prime(t) = (f^\prime)^{-1}(t)$,
\begin{align}
    T^\diamond_\eta(\mathbf{x},j) = f^\prime \Biggl( \left( 1 - \sum_{i = 1}^K e_i \right)p_{Y|X}(j|\mathbf{x}) + e_j \Biggr),
\end{align}
where $T^\diamond_\eta(\mathbf{x},j)$ is the optimal neural network learned at convergence.
Therefore, the posterior estimator obtained in the presence of label noise reads as
\begin{align}
    \hat{p}^\eta_{Y|X}(j|\mathbf{x}) = (f^*)^\prime(T^\diamond_\eta(\mathbf{x},j)) = \left( 1 - \sum_{i = 1}^K e_i \right)p_{Y|X}(j|\mathbf{x}) + e_j .
\end{align}
\end{proof}


\subsection{Proof of Theorem \ref{thm:robustness}}

\begin{appthm}{\ref{thm:robustness}}
    In a multi-class classification task, $f$-PML is noise tolerant under symmetric label noise if $\eta < \frac{K-1}{K}$.
\end{appthm}
\begin{proof}
This proof is a direct consequence of Theorem \ref{thm:multiclass_posterior}, as symmetric label noise is a particular case of asymmetric uniform off-diagonal label noise. An alternative proof could follow the same reasoning showed in \cite{priebe2022deep}. \\
The class prediction is computed as the argmax of the posterior estimate, implying that the class choice is deterministic given the posterior estimate. Let $\hat{p}_{Y|X}(y_\mathbf{x}|\mathbf{x})$ be the posterior estimator in the absence of label noise, and $\hat{p}^\eta_{Y|X}(y_\mathbf{x}|\mathbf{x})$ the posterior estimator in the presence of label noise.
Therefore, if
\begin{align}
    \hat{y}_\mathbf{x} = \argmax_{{y}_\mathbf{x} \in \mathcal{A}_y} \hat{p}_{Y|X}({y}_\mathbf{x}|\mathbf{x}) = \argmax_{{y}_\mathbf{x} \in \mathcal{A}_y} \hat{p}^\eta_{Y|X}({y}_\mathbf{x}|\mathbf{x}) = \hat{y}^\eta_\mathbf{x},
\end{align}
the class prediction and the probability of correct classification will be the same for the clean and noisy settings. Thus, we have to prove that, under symmetric noise, the argmax of the posterior estimator trained with label noise is equal to the argmax of the posterior estimator trained with the clean dataset.\\
From \eqref{eq:posterior_multiclass}, 
\begin{align}
    \hat{p}^\eta_{Y|X}(i|\mathbf{x}) = \left(1 - \sum_{j=1}^K e_j \right)\hat{p}_{Y|X}(i|\mathbf{x}) + e_i ,
\end{align}
because at convergence the posterior estimator trained over the clean dataset coincides with the true posterior. The symmetric noise scenario implies
\begin{align}
    \hat{p}^\eta_{Y|X}(i|\mathbf{x}) = \left(1 - \frac{K}{K-1}\eta\right)p_{Y|X}(i|\mathbf{x}) + \frac{\eta}{K-1}
\end{align}
when $e_i = \frac{\eta}{K-1}$.
When $\eta < \frac{K-1}{K}$, the multiplicative constant and the addition of $\frac{\eta}{K-1}$ to all the components of $p_{Y|X}$ does not modify the argmax of $p_{Y|X}$. The theorem's claim follows.
\end{proof}

\subsection{Proof of Theorem \ref{thm:bound_bias}}

\begin{appthm}{\ref{thm:bound_bias}}
    Let $T^{{(i)}}_\eta$ be the neural network at the $i$-th step of training maximizing $\mathcal{J}^\eta_f(T)$. Assume $T^{{(i)}}_\eta$ belongs to the neighborhood of $T^{\diamond}_\eta$. The bias during training is bounded as
    \begin{align}
    |p^{\diamond}_\eta - p^{{(i)}}_\eta| \leq ||(T^{\diamond}_\eta - T^{{(i)}}_\eta)||_2 ||(f^*)^{\prime \prime}(T^{{(i)}}_\eta)||_2 .
\end{align}
\end{appthm}
\begin{proof}
    The difference between $p^{\diamond}_\eta$ and $p^{{(i)}}_\eta$ can be written as
\begin{align}
    p^{\diamond}_\eta - p^{{(i)}}_\eta =& (f^*)^\prime (T_\eta^\diamond) - (f^*)^\prime(T_\eta^{(i)}) \\
    \simeq & \delta^{(i)} (f^*)^{\prime \prime}(T_\eta^{(i)}) \\
    =& (T_\eta^\diamond - T_\eta^{(i)}) (f^*)^{\prime \prime}(T_\eta^{(i)})
\end{align}
Thus,
\begin{align}
    |p^{\diamond}_\eta - p^{{(i)}}_\eta| =& |(T_\eta^\diamond - T_\eta^{(i)}) (f^*)^{\prime \prime}(T_\eta^{(i)})| \leq ||(T_\eta^\diamond - T_\eta^{(i)})||_2 ||(f^*)^{\prime \prime}(T_\eta^{(i)})||_2 
\end{align}
for the Cauchy-Schwarz inequality.
\end{proof}

\subsection{Proof of Theorem \ref{thm:convergence_multiclass_estimator}}
\begin{appthm}{\ref{thm:convergence_multiclass_estimator}}
    Let $T_{\eta j}^{\diamond}$ and $T_{\eta j}^{(i)}$ the $j$-th output of the posterior estimator at convergence and at the $i$-th iteration of training, respectively. The difference between the optimal posterior estimate without label noise and the estimate at $i$-th iteration in the presence of label noise reads as
    \begin{align}
        p_j^\diamond - p_{\eta j}^{{(i)}} \simeq \left(\sum_{n=1}^Ke_n\right) p_j^\diamond - e_j + \delta_j^{(i)}(f^*)^{\prime \prime}(T_{\eta j}^{\diamond} - \delta_j^{(i)}),
    \end{align}
    where $\delta_j^{(i)} = T_{\eta j}^{\diamond} - T_{\eta j}^{(i)}$. 
\end{appthm}
\begin{proof}
    We can study the bias of the estimator during training as
    \begin{align}
        p^\diamond - p_\eta^{(i)} =& (f^*)^\prime (T^\diamond) - (f^*)^\prime(T_\eta^{(i)}) \\
        =& (f^*)^\prime (T^\diamond) - (f^*)^\prime(T_\eta^{\diamond} - \delta^{(i)}) \\
        \simeq & (f^*)^\prime (T^\diamond) - (f^*)^\prime(T_\eta^{\diamond}) + \delta^{(i)}(f^*)^{\prime \prime}(T_\eta^\diamond - \delta^{(i)})
    \end{align}
    where the last step is obtained using the Taylor expansion.
    In the binary case, for the $j$-th class, we get
    \begin{align}
        p_j^\diamond - p_{\eta j}^{(i)} \simeq & (f^*)^\prime (T_j^\diamond) - [(1-e_0 -e_1)(f^*)^\prime(T_j^\diamond) + e_j] + \delta_j^{(i)}(f^*)^{\prime \prime}(T_{\eta j}^\diamond - \delta_j^{(i)}) \\
        =& (f^*)^\prime(T_j^\diamond)[1-(1-e_0-e_1)] - e_j + \delta_j^{(i)}(f^*)^{\prime \prime}(T_{\eta j}^\diamond - \delta_j^{(i)}) \\
        =& [e_0+e_1] (f^*)^\prime(T_j^\diamond) - e_j + \delta_j^{(i)}(f^*)^{\prime \prime}(T_{\eta j}^\diamond - \delta_j^{(i)}) \\
        =& [e_0+e_1] p_j^\diamond - e_j + \delta_j^{(i)}(f^*)^{\prime \prime}(T_{\eta_j}^\diamond - \delta_j^{(i)}).
    \end{align}
    
    In the multi-class case, for the $j$-th output of the discriminator, we get 
    \begin{align}
        p_j^\diamond - p_{\eta j}^{(i)} \simeq & (f^*)^\prime (T_j^\diamond) - [(1- \sum_{i=1}^K e_i)(f^*)^\prime(T_j^\diamond) + e_j] + \delta_j^{(i)}(f^*)^{\prime \prime}(T_{\eta j}^\diamond - \delta_j^{(i)}) \\
        =& \left(\sum_{i=1}^Ke_i\right) (f^*)^\prime(T_j^\diamond) - e_j + \delta_j^{(i)}(f^*)^{\prime \prime}(T_{\eta j}^\diamond - \delta_j^{(i)}) \\
        =& \left(\sum_{i=1}^Ke_i\right) p_j^\diamond - e_j + \delta_j^{(i)}(f^*)^{\prime \prime}(T_{\eta j}^\diamond - \delta_j^{(i)}) .
    \end{align}
\end{proof}



\section{Comparison with Related Work}
\label{sec:appendix_related}

\subsection{Active Passive Losses}
\label{subsec:APLs}

In this section, we first recall the definitions of active and passive losses from \cite{ma2020normalized}. Then, we show that the class of objective functions in \eqref{eq:supervised_general_value_function} is composed by the sum of an active and a passive objective functions.

\begin{definition}[Active loss function (see \cite{ma2020normalized})]
\label{def:active_loss}
    $\mathcal{J}_{Active}$ is an active loss function if $\forall(\mathbf{x},\mathbf{y}_\mathbf{x}) \in \mathcal{D}$, $\forall k \neq \mathbf{y}_\mathbf{x}$ $l(f(\mathbf{x}),k) = 0$.
\end{definition}

\begin{definition}[Passive loss function (see \cite{ma2020normalized})]
\label{def:passive_loss}
    $\mathcal{J}_{Passive}$ is a passive loss function if $\forall(\mathbf{x},\mathbf{y}_\mathbf{x}) \in \mathcal{D}$, $\exists k \neq \mathbf{y}_\mathbf{x}$ such that $l(f(\mathbf{x}),k) \neq 0$.
\end{definition}
Definition \ref{def:active_loss} describes objective functions that are only affected by the prediction corresponding to the label. All the predictions corresponding to a class different from the label of the sample $\mathbf{x}$ are irrelevant. 
Definition \ref{def:passive_loss} describes objective functions for which at least one of the neural network's predictions corresponding to a class different from the label contributes to the objective function value.

Following definitions \ref{def:active_loss} and \ref{def:passive_loss}, the class of objective functions in \eqref{eq:supervised_general_value_function} can be rewritten as $\mathcal{J}_f = \mathcal{J}_{Active} + \mathcal{J}_{Passive}$. 


In \cite{ye2023active}, the authors study the APLs proposed in \cite{ma2020normalized} and notice that the passive losses proposed in \cite{ma2020normalized} are all scaled versions of MAE. Therefore, they propose a new class of passive loss functions based on complementary label learning and vertical flipping. They show that this new class of passive losses perform better than the one used in \cite{ma2020normalized}. 

Differently from \cite{ye2023active}, in this paper the active and passive objective functions are directly related to the $f$-divergence used and therefore the passive term depends on the active. In other words, while APLs and ANLs are the sum of their parts, the objective functions of $f$-PML are greater than the sum of their parts. 


\subsection{$f$-Divergence for Noisy Labels}

The $f$-divergence has been used in learning with noisy labels in \cite{wei2020optimizing}, where the authors maximize the $f$-MI (which is a generalization of the MI using the $f$-divergence) between the label distribution and the classifier's output distribution. 
Several machine learning approaches rely on the maximization of MI, for instance for representation learning \cite{hjelm2018learning} and communication engineering \cite{letizia2023cooperative} applications. However, the maximization of MI does not always lead to learning the best models, as showed in \cite{tschannen2019mutual} for the representation learning domain. 
In this specific scenario, there is no guarantee that the maximization of the $f$-MI is a classification objective which leads to a Bayes classifier
\begin{align}
\label{eq:bayes_optimal_classifier}
    C^{B}(\mathbf{x}) = \argmax_{i\in \{1,\dots, K\}} P(Y=i|X=\mathbf{x}).
\end{align}
The authors of \cite{wei2020optimizing}, in fact, proved that in the binary classification scenario maximizing the $f$-MI leads to the Bayes optimal classifier only when the classes in the dataset have equal prior probability (i.e., it is a balanced dataset) and when using a restricted set of $f$-divergences (e.g., the total variation). They extend their findings for the multi-class scenario only for confident classifiers. \\
Differently, the maximization of the \textbf{PMI} between images and corresponding labels on which $f$-PML relies corresponds to the solution of the optimal classification approach under a Bayesian setting \cite{tonello2022mind}, which is the MAP approach, returning the Bayes optimal classifier \eqref{eq:bayes_optimal_classifier} by definition.

In addition, variational MI estimators are upper bounded \cite{mcallester2020formal}. The main reason is that they need to draw samples from $p_X(\mathbf{x})p_Y(y)$. However, practically it is difficult to ensure that, given a batch of samples drawn from $p_{XY}(\mathbf{x}, y_\mathbf{x})$, a random shuffle/derangement of the batch of $y$ returns a batch of samples from $p_X(\mathbf{x})p_Y(y)$. This is still an open problem \cite{mcallester2020formal, letizia2023variational} which bounds MI estimates. 
Differently, $f$-PML does not need to break the relationship between the realizations of $X$ and $Y$ through a shuffling mechanism to draw the samples from $p_X(\mathbf{x})p_Y(y)$, because it only needs samples from $p_{XY}(\mathbf{x},y_\mathbf{x})$.

Finally, the objective function in \cite{wei2020optimizing} is robust to symmetric and asymmetric off-diagonal label noise for a restricted class of $f$-divergences, while $f$-PML is robust to symmetric label noise for any $f$-divergence.

\section{Additional Experimental Results}
\label{sec:appendix_results}

\subsection{Implementation Details}
\label{subsec:appendix_implementation_details}

\paragraph{Datasets description} 
For the binary classification scenario, we use the breast cancer dataset \cite{breast_cancer_wisconsin} available on Scikit-learn \cite{pedregosa2011scikit}. It contains $569$ samples and $30$ features.
For the multiclass classification task, we use datasets with synthetic label noise generated from CIFAR-10 and CIFAR-100 \cite{krizhevsky2009learning}. These consist of $60k$ $32\times32$ images split in $50k$ for training and $10k$ for test. CIFAR-10 contains $10$ classes, with $6000$ images per class. CIFAR-100 contains $100$ classes, with $600$ images per class. Following previous work, the synthetic symmetric label noise is generated by randomly flipping the label of a given percentage of samples into a fake label with a uniform probability, while the asymmetric label noise is generated by flipping labels for specific classes. For the uniform off-diagonal label noise, we use a custom transition matrix which is defined in Sec. \ref{subsubsec:appendix_experiments_correction}.
For datasets with realistic label noise, we use CIFAR-10N and CIFAR-100N \cite{wei2021learning}. CIFAR-10N contains human annotations from three independent workers (Random 1, Random 2, and Random 3) which are combined by majority voting to get an aggregated label (Aggregate) and to get wrong labels (Worst). CIFAR-100N contains human annotations submitted for the fine classes.

\paragraph{Hyperparameters and network architecture} 
We use a ResNet34 \cite{resnet} for almost all the experiments of $f$-PML, consistently with the literature. For the comparisons with APL-like objective functions, we use the same $8$-layer CNN used in Ma et al. \cite{ma2020normalized, ye2023active}. For $f$-PML$_{\text{Pro}}$, we use the Promix architecture, consisting of $2$ ResNet18.
Optimization is executed using SGD with a momentum of $0.9$. The learning rate is initially set to $0.02$ and a cosine annealing scheduler \cite{loshchilov2016sgdr} decays it during training. For the ProMix training strategy and architecture, we use the same hyperparameters reported in \cite{wang2022promix}\footnote{See the GitHub repository of ProMix \url{https://github.com/Justherozen/ProMix}}. For the experiments on the binary dataset, we trained the models for $100$ epochs, with a batch size of $32$. For the comparison with APL-like losses on the CIFAR-10 dataset, we trained the neural networks for $120$ epochs, with a batch size of $128$. For any other dataset and scenario, we trained $f$-PML for $300$ epochs, and $f$-PML$_{Pro}$ for $600$ epochs, with a batch size of $128$ and $256$, respectively. 
For $f$-PML$_{Pro}$ and ProMix$^*$, we use the same hyperparameters reported in \cite{wang2022promix}. All the tables report the mean over $5$ independent runs of the code with different random seeds. Some also report the standard deviation. 

\paragraph{Baselines} All the baselines are reported in the following: standard cross-entropy minimization approach (CE), Forward \cite{patrini2017making}, GCE \cite{zhang2018generalized}, Co-teaching \cite{han2018co}, Co-teaching+ \cite{yu2019does}, SCE \cite{wang2019symmetric}, NLNL \cite{kim2019nlnl}, JoCoR \cite{wei2020combating}, ELR \cite{liu2020early}, Peer Loss \cite{liu2020peer}, NCE+RCE/NCE+MAE/NFL+RCE/NFL+MAE \cite{ma2020normalized}, NCE+AEL/NCE+AGCE/NCE+AUL \cite{zhou2021asymmetric}, F-Div \cite{wei2020optimizing}, Divide-Mix \cite{li2020dividemix}, Negative-LS \cite{wei2022smooth}, CORES$^2$ \cite{cheng2021learning}, SOP \cite{liu2022robust}, ProMix \cite{wang2022promix}, ANL-CE/ANL-FL \cite{ye2023active}, RDA \cite{lienen2024mitigating}, SGN \cite{englesson2024robust}.

\subsection{Additional Results}


\subsubsection{Correction Methods}
\label{subsubsec:appendix_experiments_correction}
For a multiclass classification problem with $K$ classes, the transition matrix used is defined as
\begin{align}
    T = \begin{bmatrix}
    P(Y_\eta = 1|Y=1)  & \cdots & P(Y_\eta=K|Y=1)\\
    \vdots & \ddots & \vdots \\
    P(Y_\eta=1|Y=K) & \cdots & P(Y_\eta =K|Y=K) \\
    \end{bmatrix}.
\end{align}
For the experimental part of the paper, the transition matrices are defined as: 
\begin{itemize}
    \item Binary cancer dataset 
    \begin{align}
        T = \begin{bmatrix}
            1 - e_1 & e_1 \\
            e_0 & 1 - e_0 
        \end{bmatrix},
    \end{align}
    where $e_0, e_1$ are specified for each specific example.
    \item CIFAR10, uniform off-diagonal low noise matrix \cite{wei2020combating}
    \begin{align}
    T = \begin{bmatrix}
    0.82 & 0.03 & 0.01 & 0.023 & 0.017 & 0.022 & 0.021 & 0.018 & 0.019 & 0.02\\
    0.02 & 0.83 & 0.01 & 0.023 & 0.017 & 0.022 & 0.021 & 0.018 & 0.019 & 0.02 \\
    0.02 & 0.03 & 0.81 & 0.023 & 0.017 & 0.022 & 0.021 & 0.018 & 0.019 & 0.02 \\
    0.02 & 0.03 & 0.01 & 0.823 & 0.017 & 0.022 & 0.021 & 0.018 & 0.019 & 0.02 \\
    0.02 & 0.03 & 0.01 & 0.023 & 0.817 & 0.022 & 0.021 & 0.018 & 0.019 & 0.02 \\
    0.02 & 0.03 & 0.01 & 0.023 & 0.017 & 0.822 & 0.021 & 0.018 & 0.019 & 0.02 \\
    0.02 & 0.03 & 0.01 & 0.023 & 0.017 & 0.022 & 0.821 & 0.018 & 0.019 & 0.02 \\
    0.02 & 0.03 & 0.01 & 0.023 & 0.017 & 0.022 & 0.021 & 0.818 & 0.019 & 0.02 \\
    0.02 & 0.03 & 0.01 & 0.023 & 0.017 & 0.022 & 0.021 & 0.018 & 0.819 & 0.02 \\
    0.02 & 0.03 & 0.01 & 0.023 & 0.017 & 0.022 & 0.021 & 0.018 & 0.019 & 0.82 \\
    \end{bmatrix}
    \end{align}
    \item CIFAR10, uniform off-diagonal high noise matrix \cite{wei2020combating}
    \begin{align}
    T = \begin{bmatrix}
    0.46 & 0.07 & 0.04 & 0.05 & 0.06 & 0.04 & 0.06 & 0.07 & 0.08 & 0.07\\
    0.05 & 0.48 & 0.04 & 0.05 & 0.06 & 0.04 & 0.06 & 0.07 & 0.08 & 0.07\\
    0.05 & 0.07 & 0.45 & 0.05 & 0.06 & 0.04 & 0.06 & 0.07 & 0.08 & 0.07\\
    0.05 & 0.07 & 0.04 & 0.46 & 0.06 & 0.04 & 0.06 & 0.07 & 0.08 & 0.07\\
    0.05 & 0.07 & 0.04 & 0.05 & 0.47 & 0.04 & 0.06 & 0.07 & 0.08 & 0.07\\
    0.05 & 0.07 & 0.04 & 0.05 & 0.06 & 0.45 & 0.06 & 0.07 & 0.08 & 0.07\\
    0.05 & 0.07 & 0.04 & 0.05 & 0.06 & 0.04 & 0.47 & 0.07 & 0.08 & 0.07\\
    0.05 & 0.07 & 0.04 & 0.05 & 0.06 & 0.04 & 0.06 & 0.48 & 0.08 & 0.07\\
    0.05 & 0.07 & 0.04 & 0.05 & 0.06 & 0.04 & 0.06 & 0.07 & 0.49 & 0.07\\
    0.05 & 0.07 & 0.04 & 0.05 & 0.06 & 0.04 & 0.06 & 0.07 & 0.08 & 0.48\\
    \end{bmatrix}.
    \end{align}
\end{itemize}

\subsubsection{Additional Numerical Results}
\label{subsec:appendix_image_datasets}


\paragraph{Objective function and posterior correction} Table \ref{tab:breast_classification_0_2__0_4} shows the comparison between KL-PML, SL-PML, and JS-PML in the absence and presence of binary label noise. With label noise, we compare $f$-PML without correction, with posterior correction, and with objective function correction, for $e_0=0.2$ and $e_1=0.4$. 
\begin{table}[h]
\caption{Test accuracy for the breast cancer test dataset for $[e_0, e_1] = [0.2, 0.4]$.} 
  \begin{center}
  \begin{small}
  \begin{sc}
    \begin{tabular}{ c c c c | c } 
     \hline
     Div. & No Cor. & P. Cor. & O.F. Corr. & No Noise \\
     \hline
     KL-PML & $90.4$ & $92.2$ & $94.7$ & $98.2$ \\
     SL-PML & $87.7$ & $91.3$ & $93.9$ & $98.2$ \\
     JS-PML & $89.0$ & $92.2$ & $94.7$ & $98.2$ \\
     \hline
    \end{tabular}
    \end{sc}
    \end{small}
    \end{center}
    \vskip -0.1in
    \label{tab:breast_classification_0_2__0_4}
\end{table}

\paragraph{Additional experimental results} In this paragraph, we compare the test accuracies for asymmetric label noise for the objective functions that have an APL-like formulation and for other methods that only propose objective functions\footnote{The result of ANLs was obtained by including an L1 regularization loss in the objective function}, without using refined training strategies or complex architectures. 
The acronyms in Tabs. \ref{tab:CIFAR10-100_asymm_appendix}, \ref{tab:CIFAR10-100_asymm} are the following: Reverse Cross Entropy (RCE), Focal Loss (FL), Asymmetric Generalized Cross Entropy (AGCE), Asymmetric Unhinged Loss (AUL), and Asymmetric Exponential Loss (AEL) (the last three have been proposed in \cite{zhou2021asymmetric}). For CIFAR-100, in Tab. \ref{tab:CIFAR10-100_asymm_appendix}, we used the same change of variable proposed in Novello \& Tonello \cite{pmlr-v235-novello24a}. 
\begin{table}[h]
\caption{Test accuracy achieved on CIFAR-10 and CIFAR-100 with asymmetric noise. An 8-layer CNN is used for CIFAR-10. The ResNet34 is used for CIFAR-100.} 
  \begin{center}
  \begin{small}
  \begin{sc}
    \begin{tabular}{ c c c c c c c } 
    \toprule
    \textbf{Method} & \multicolumn{3}{c}{\textbf{CIFAR-10}} & \multicolumn{3}{c}{\textbf{CIFAR-100}} \\
    \cmidrule{2-7} 
     & $20\%$ & $30\%$& $40\%$ & $20\%$ & $30\%$ & $40\%$ \\
     \toprule
     CE & $83.00_{\pm 0.33}$ & $78.15_{\pm 0.17}$ & $73.69_{\pm 0.20}$ & $58.25_{\pm 1.00}$ & $50.30_{\pm 0.19}$ & $41.53_{\pm 0.34}$ \\
     MAE & $79.63_{\pm 0.74}$ & $67.35_{\pm 3.41}$ & $57.36_{\pm 2.37}$ & $6.19_{\pm 0.42}$ & $5.82_{\pm 0.96}$ & $3.96_{\pm 0.35}$ \\
     GCE & $85.55_{\pm 0.24}$ & $79.32_{\pm 0.52}$ & $72.83_{\pm 0.17}$ & $59.06_{\pm 0.46}$ & $53.88_{\pm 0.96}$ & $41.51_{\pm 0.52}$\\
     SCE & $86.22_{\pm 0.44}$ & $80.20_{\pm 0.20}$ & $74.01_{\pm 0.52}$ & $57.78_{\pm 0.83}$ & $50.15_{\pm 0.12}$ & $41.33_{\pm 0.86}$ \\
     NLNL & $84.74_{\pm 0.08}$ & $81.26_{\pm 0.43}$ & $76.97_{\pm 0.52}$ & $50.19_{\pm 0.56}$ & $42.81_{\pm 1.13}$ & $35.10_{\pm 0.20}$ \\
     NCE+RCE & $88.36_{\pm 0.13}$ & $84.84_{\pm 0.16}$ & $77.75_{\pm 0.37}$ & $62.77_{\pm 0.53}$ & $55.62_{\pm 0.56}$ & $42.46_{\pm 0.42}$ \\
     NCE+AGCE & $88.48_{\pm 0.09}$ & $84.79_{\pm 0.15}$ & $78.60_{\pm 0.41}$ & $64.05_{\pm 0.25}$ & $56.36_{\pm 0.59}$ & $44.90_{\pm 0.62}$ \\
     ANL-CE & $89.13_{\pm 0.11}$ & $85.52_{\pm 0.24}$ & $77.63_{\pm 0.31}$ & $66.27_{\pm 0.19}$ & $59.76_{\pm 0.34}$ & $45.41_{\pm 0.68}$\\
     ANL-FL & $89.09_{\pm 0.31}$ & $85.81_{\pm 0.23}$ & $77.73_{\pm 0.31}$ & $66.26_{\pm 0.44}$ & $59.68_{\pm 0.86}$ & $46.65_{\pm 0.04}$ \\
     \hline
     SL-PML & $\textbf{89.14}_{\pm 0.12}$ & $\textbf{86.67}_{\pm 0.27}$ & $63.12_{\pm 0.48}$ & $70.90_{\pm 39}$ & $67.36_{\pm 0.74}$ & $64.59_{\pm 0.98}$ \\
     GAN-PML & $89.02_{\pm 0.10}$ & $86.14_{\pm 0.21}$ & $\textbf{82.15}_{\pm 0.34}$ & $\textbf{73.58}_{\pm 0.41}$ & $\textbf{69.80}_{\pm 0.92}$ & $\textbf{65.93}_{\pm 0.95}$\\
     \hline
    \end{tabular}
    \end{sc}
    \end{small}
    \end{center}
    \vskip -0.1in
    \label{tab:CIFAR10-100_asymm_appendix}
\end{table}

Training the ResNet50 on ImageNet, we obtain a slightly lower accuracy, but still higher than other approaches that train on WebVision mini and then test on the subset of ImageNet of the same classes (GAN-PML obtains $\textbf{68.88}$).

\end{document}